\documentclass[final]{cvpr}

\usepackage{times}
\usepackage{epsfig}
\usepackage{graphicx}
\usepackage{amsmath}
\usepackage{amssymb}
\usepackage{multirow}
\usepackage{subcaption}
\usepackage[dvipsnames]{xcolor}
\usepackage{tabularx}
\usepackage{booktabs}
\usepackage[font=small]{caption}
\usepackage[pagebackref=true,breaklinks=true,colorlinks,bookmarks=false]{hyperref}

\def\cvprPaperID{10509} 
\def\confYear{CVPR 2021}

\begin{document}

\title{Real-Time High-Resolution Background Matting}

\author{
Shanchuan Lin\thanks{Equal contribution.}\hspace{7mm}
Andrey Ryabtsev\textsuperscript{*}\hspace{7mm}
Soumyadip Sengupta\\
Brian Curless\hspace{7mm}
Steve Seitz\hspace{7mm}
Ira Kemelmacher-Shlizerman \\

University of Washington \\
{\tt\small \{linsh,ryabtsev,soumya91,curless,seitz,kemelmi\}@cs.washington.edu}
}

\maketitle

\begin{abstract}
\vspace{-0.5em}
We introduce a real-time, high-resolution background replacement technique which operates at 30fps in 4K resolution, and 60fps for HD on a modern GPU. Our technique is based on background matting, where an additional frame of the background is captured and used in recovering the alpha matte and the foreground layer. The main challenge is to compute a high-quality alpha matte, preserving strand-level hair details, while processing high-resolution images in real-time. To achieve this goal, we employ two neural networks; a base network computes a low-resolution result which is refined by a second network operating at high-resolution on selective patches. We introduce two large-scale video and image matting datasets:  {\em VideoMatte240K} and {\em PhotoMatte13K/85}. Our approach yields higher quality results compared to the previous state-of-the-art in background matting, while simultaneously yielding a dramatic boost in both speed and resolution. Our code and data is available at \small{\url{https://grail.cs.washington.edu/projects/background-matting-v2/}}
\end{abstract}
\vspace{-1em}
\section{Introduction}
\label{sec:intro}

Background replacement, a mainstay in movie special effects, now enjoys wide-spread use in video conferencing tools like Zoom, Google Meet, and Microsoft Teams.  In addition to adding entertainment value, background replacement can enhance privacy, particularly in situations where a user may not want to share details of their location and environment to others on the call.  A key challenge of this video conferencing application is that users do not typically have access to a green screen or other physical props used to facilitate background replacement in movie special effects.

While many tools now provide background replacement functionality, they yield artifacts at boundaries, particularly in areas where there is fine detail like hair or glasses (Figure 1).  In contrast, traditional image matting methods \cite{chen2013knn,levin2007closed,levin2008spectral,sun2004poisson,gastal2010shared,aksoy2017designing,chuang2001bayesian} provide much higher quality results, but do not run in real-time, at high resolution, and frequently require manual input.  
In this paper, we introduce the first fully-automated, real-time, high-resolution matting technique, producing state-of-the-art results at 4K (3840$\times$2160) at 30fps and HD (1920$\times$1080) at 60fps. Our method relies on capturing an extra background image to compute the alpha matte and the foreground layer, an approach known as background matting.

Designing a neural network that can achieve real-time matting on high-resolution videos of people is extremely challenging, especially when fine-grained details like strands of hair are important;  in contrast, the previous state-of-the-art method \cite{sengupta2020background} is limited to 512$\times$512 at 8fps. Training a deep network on such a large resolution is extremely slow and memory intensive. It also requires large volumes of images with high-quality alpha mattes to generalize; the publicly available datasets \cite{xu2017deep,qiao2020attention} are too limited.

Since it is difficult to collect a high-quality dataset with manually curated alpha mattes in large quantities, we propose to train our network with a series of datasets, each with different characteristics. To this end, we introduce VideoMatte240K and PhotoMatte13K/85 with high-resolution alpha mattes and foreground layers extracted with chroma-key software. We first train our network on these larger databases of alpha mattes with significant diversity in human poses to learn robust priors.  We then train on publicly available datasets \cite{xu2017deep,qiao2020attention} that are manually curated to learn fine-grained details.

To design a network that can handle high-resolution images in real-time, we observe that relatively few regions in the image require fine-grained refinement. Therefore, we introduce a base network that predicts the alpha matte and foreground layer at lower resolution along with an error prediction map which specifies areas that may need high-resolution refinement. A refinement network then takes the low-resolution result and the original image to generate high-resolution output only at select regions.

We produce state-of-the-art background matting results in real-time on challenging real-world videos and images of people. We will release our VideoMatte240K and PhotoMatte85 datasets and our model implementation. 

\section{Related Work}
\label{sec:related}

Background replacement can be achieved with segmentation or matting. While binary segmentation is fast and efficient, the resulting composites have objectionable artifacts. Alpha matting can produce visually pleasing composites but often requires either manual annotations or a known background image. In this section, we discuss related works that perform background replacement with segmentation or matting.

\textbf{Segmentation.} The literature in both instance and semantic segmentation is vast and out of scope for this paper, so we will review the most relevant works. Mask RCNN \cite{he2017mask} is still a top choice for instance segmentation while DeepLabV3+ \cite{deeplabv3plus2018} is a state-of-the-art semantic segmentation network. We incorporate the Atrous Spatial Pyramid Pooling (ASPP) module from DeepLabV3 \cite{chen2017rethinking} and DeepLabV3+ within our network. Since segmentation algorithms tend to produce coarse boundaries especially at higher resolutions,  Kirillov \textit{et al.} presented PointRend \cite{kirillov2020pointrend} which samples points near the boundary and iteratively refines the segmentation. This produces high-quality segmentation for large image resolutions with significantly cheaper memory and computation. Our method adopts this idea to the matting domain via learned refinement-region selection and a convolutional refinement architecture that improves the receptive field. Specific applications of human segmentation and parsing have also received considerable attention in recent works \cite{zhang2019pose2seg,liang2018look}.

\textbf{Trimap-based matting.} Traditional (non-learning based) matting algorithms \cite{chen2013knn,levin2007closed,levin2008spectral,sun2004poisson,gastal2010shared,aksoy2017designing,chuang2001bayesian} require manual annotation (a trimap) and solve for the alpha matte in the `unknown' region of the trimap. Different matting techniques are reviewed in the survey by Wang and Cohen \cite{wang2008image}. Xu \textit{et al.} \cite{xu2017deep} introduced a matting dataset and used a deep network with a trimap input to predict the alpha matte. Many recent approaches rely on this dataset to learn matting, e.g., Context-Aware Matting \cite{hou2019context}, Index Matting \cite{lu2019indices}, sampling-based matting \cite{Tang_2019_CVPR} and opacity propagation-based matting \cite{li2020hierarchical}. Although the performance of these methods depends on the quality of the annotations, some recent methods consider coarse \cite{liu2020boosting} or faulty human annotations \cite{cai2019disentangled} to predict the alpha matte.

\textbf{Matting without any external input.} Recent approaches have also focused on matting humans without any external input. Portrait matting without a trimap \cite{zhu2017fast,shen2016deep} is one of the more successful applications due to less variability among portrait images compared to full body humans. Soft segmentation for natural images had also been explored in \cite{aksoy2018semantic}. Recent approaches like Late Fusion Matting \cite{zhang2019late} and HAttMatting \cite{qiao2020attention} aim to solve for the alpha matte directly from the image, but these approaches can often fail to generalize as shown in \cite{sengupta2020background}.

\textbf{Matting with a known natural background.} Matting with known natural background had been previously explored in \cite{qian1999video}, Bayesian matting~\cite{chuang2001bayesian} and Poisson matting \cite{sun2004poisson,gong2009near} which also requires a trimap. Recently Sengupta \textit{et al.} \cite{sengupta2020background} introduced Background Matting (BGM) where an additional background image is captured and it provides a significant cue to predict the alpha matte and the foreground layer. Although this method showed high-quality matting results, the architecture is limited to 512$\times$512 resolution and runs only at 8fps. In contrast, we introduce a real-time unified matting architecture that operates on 4K videos at 30fps and HD videos at 60fps, and produces higher quality results than BGM.
\section{Our Dataset}
\label{sec:dataset}

\begin{figure}[t]
    \vspace{-1em}
    \centering
    \subcaptionbox{VideoMatte240K}{
        \includegraphics[width=0.45\textwidth]{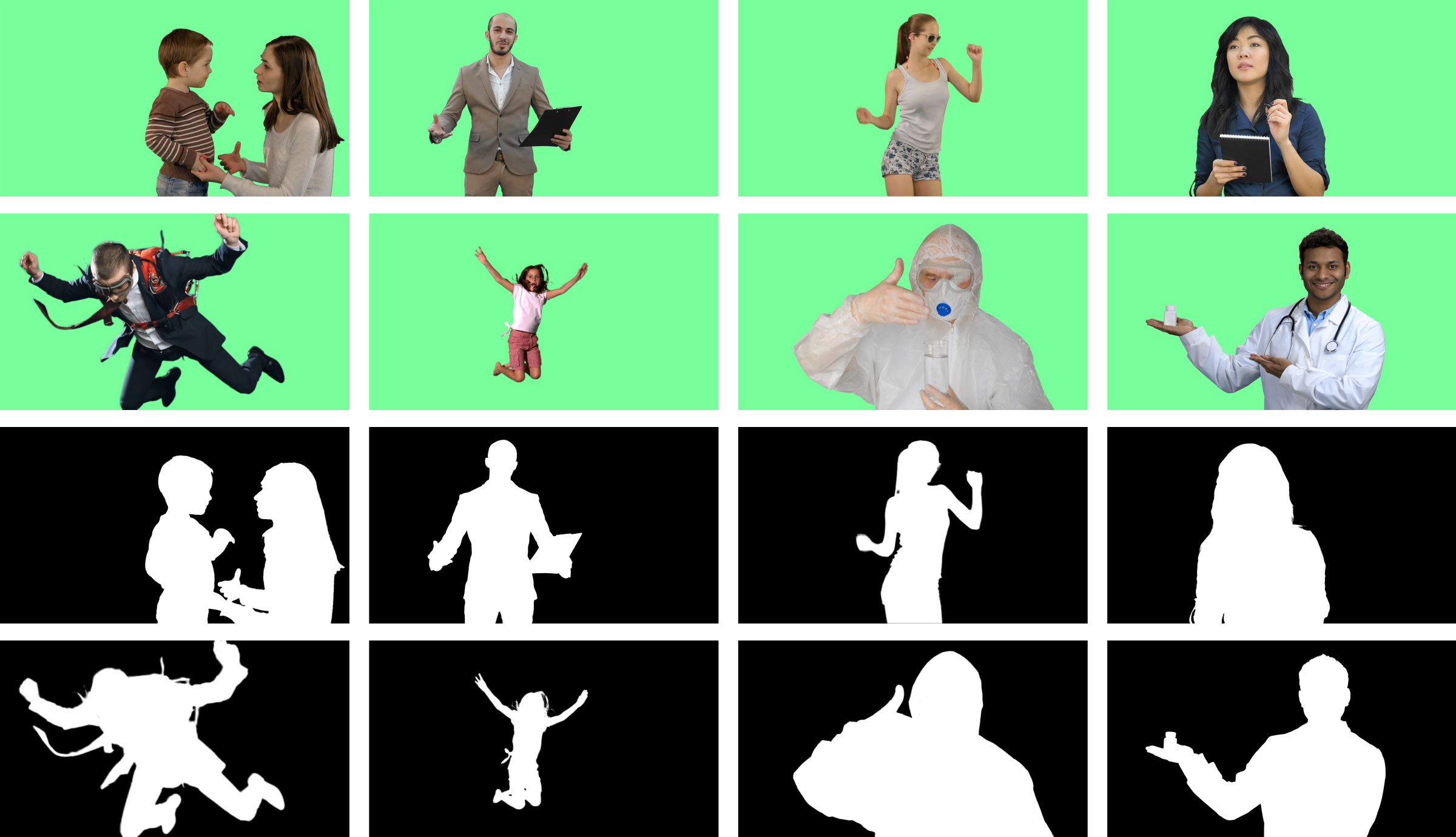}
    }
    \subcaptionbox{PhotoMatte13K/85}{
        \includegraphics[width=0.45\textwidth]{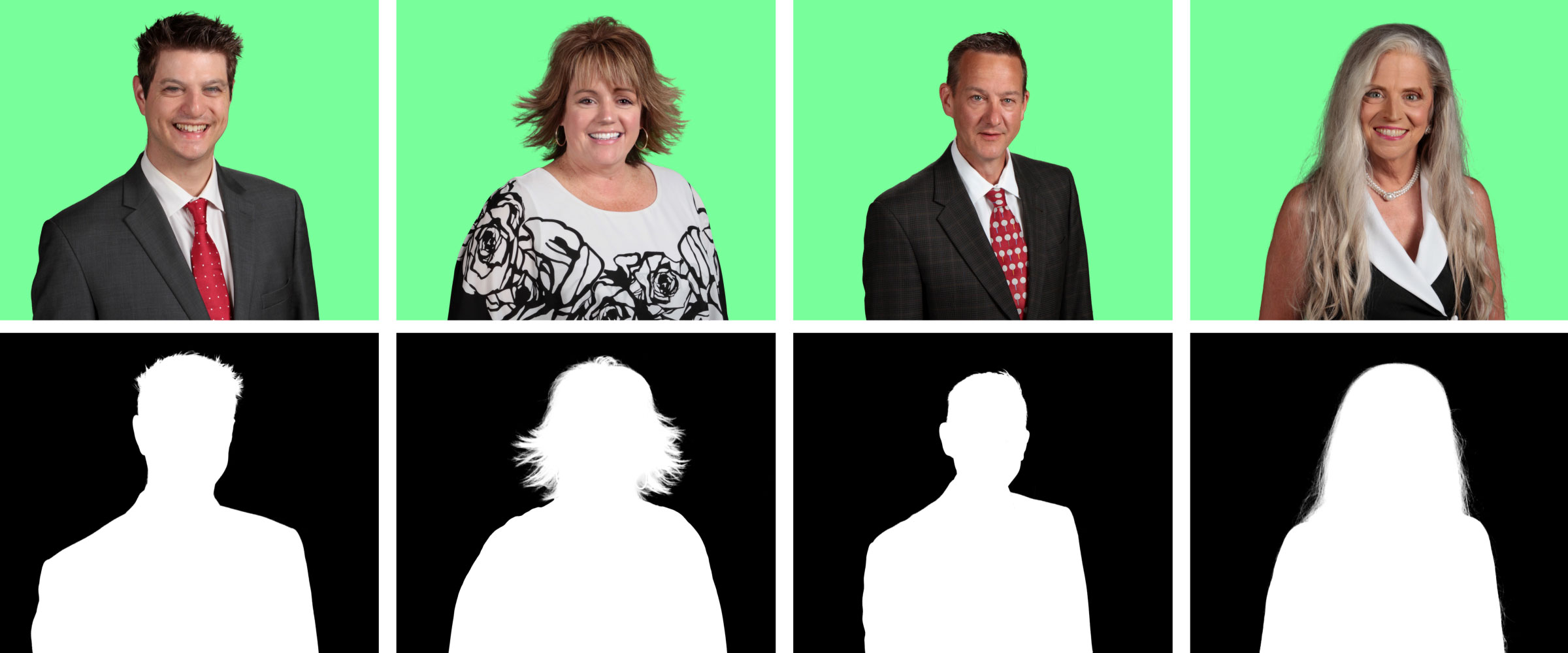}
    }
    \vspace{-0.5em}
    \caption{We introduce two large-scale matting datasets containing 240k unique frames and 13k unique photos.}
    \label{fig:dataset}
    \vspace{-1.5em}
\end{figure}

Since it is extremely difficult to obtain a large-scale, high-resolution, high-quality matting dataset where the alpha mattes are cleaned by human artists, we rely on multiple datasets including our own collections and publicly available datasets.

\textbf{Publicly available datasets.} The Adobe Image Matting (AIM) dataset ~\cite{xu2017deep} provides 269 human training samples and 11 test samples, averaging around 1000$\times$1000 resolution. We also use a humans-only subset of Distinctions-646~\cite{qiao2020attention} consisting of 362 training and 11 test samples, averaging around 1700$\times$2000 resolution. The mattes were created manually and are thus high-quality. However 631 training images are not enough to learn large variations in human poses and finer details at high resolution, so we introduce 2 additional datasets.

\textbf{VideoMatte240K.} We collect 484 high-resolution green screen videos and generate a total of 240,709 unique frames of alpha mattes and foregrounds with chroma-key software Adobe After Effects. The videos are purchased as stock footage or found as royalty-free materials online. 384 videos are at 4K resolution and 100 are in HD. We split the videos by 479 : 5 to form the train and validation sets. The dataset consists of a vast amount of human subjects, clothing, and poses that are helpful for training robust models. We are releasing the extracted alpha mattes and foregrounds as a dataset to the public. To our knowledge, our dataset is larger than all existing matting datasets publicly available by far, and it is the first public video matting dataset that contains continuous sequences of frames instead of still images, which can be used in future research to develop models that incorporate motion information. 

\textbf{PhotoMatte13K/85.} We acquired a collection of 13,665 images shot with studio-quality lighting and cameras in front of a green-screen, along with mattes extracted via chroma-key algorithms with manual tuning and error repair. We split the images by 13,165 : 500 to form the train and validation sets. These mattes contain a narrow range of poses but are high resolution, averaging around 2000$\times$2500, and include details such as individual strands of hair. We refer to this dataset as PhotoMatte13K. However privacy and licensing issues prevent us from sharing this set; thus, we also collected an additional set of 85 mattes of similar quality for use as a test set, which we are releasing to the public as PhotoMatte85. In Figure \ref{fig:dataset} we show examples from the VideoMatte240K and PhotoMatte13K/85 datasets.


We crawl 8861 high-resolution background images from Flickr and Google and split them by 8636 : 200 : 25 to use when constructing the train, validation, and test sets. We will release the test set in which all images have a CC license (see appendix for details).
\section{Our Approach}
\label{sec:model}

\begin{figure*}[t]
    \vspace{-1em}
    \centering
    \includegraphics[width=\textwidth]{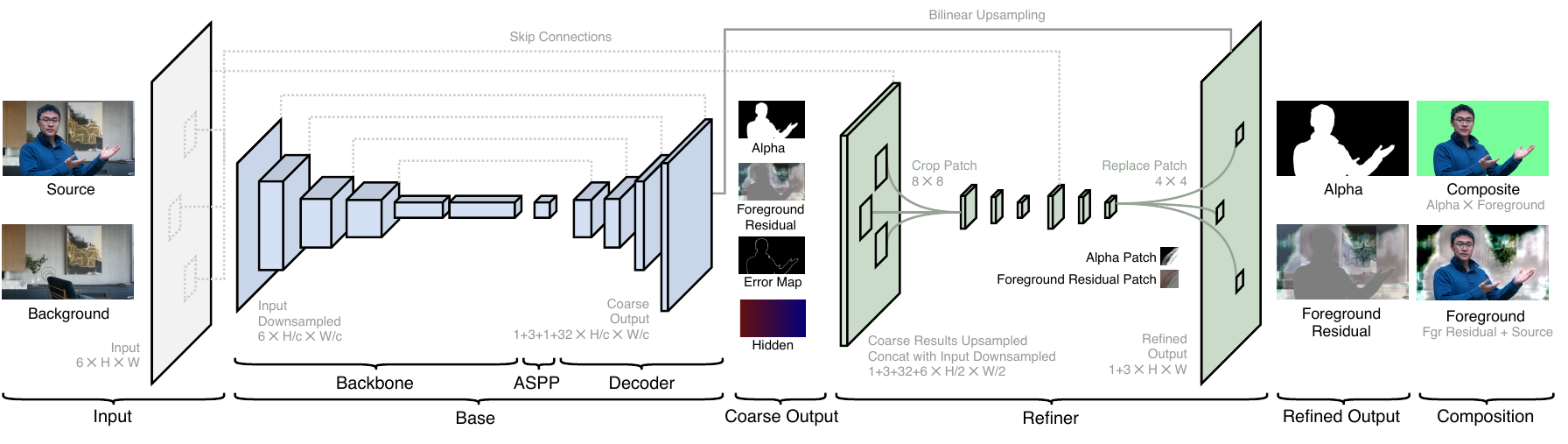}
    \vspace{-2em}
    \caption{The base network $G_{\text{base}}$ (blue) operates on the downsampled input to produce coarse-grained results and an error prediction map. The refinement network $G_{\text{refine}}$ (green) selects error-prone patches and refines them to the full resolution.}
    \label{fig:model_overview}
\end{figure*}

\begin{figure}
    \vspace{-1em}
    \centering
    \includegraphics[width=0.45\textwidth]{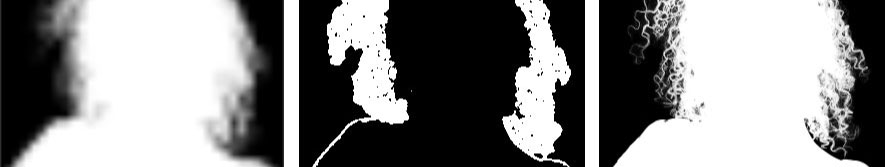}
    \newcolumntype{Y}{>{\centering\arraybackslash}X}
    \begin{small}
        \begin{tabularx}{.45\textwidth}{YYY}
             (a) Coarse & (b) Selection & (c) Refined 
        \end{tabularx}
    \end{small}
    \vspace{-0.5em}
    \caption{We only refine on error-prone regions (b) and directly upsample the rest to save computation.}
    \label{fig:model-efficient-refinement}
    \vspace{-1.5em}
\end{figure}

Given an image $I$ and the captured background $B$ we predict the alpha matte $\alpha$ and the foreground $F$, which will allow us to composite over any new background by $I'=\alpha F + (1-\alpha) B'$, where $B'$ is the new background. Instead of solving for the foreground directly, we solve for foreground residual $F^{R} = F - I$. Then, $F$ can be recovered by adding $F^{R}$ to the input image $I$ with suitable clamping: $F = \max(\min(F^{R} + I,1),0)$. We find this formulation improves learning, and allows us to apply a low-resolution foreground residual onto a high-resolution input image through upsampling, aiding our architecture as described later.

Matting at high resolution is challenging, as applying a deep network directly incurs impractical computation and memory consumption. As Figure \ref{fig:model-efficient-refinement} shows, human mattes are usually very sparse, where large areas of pixels belong to either background ($\alpha=0$) or foreground ($\alpha=1$), and only a few areas involve finer details, e.g., around the hair, glasses, and person's outline. Thus instead of designing one network that operates on high-resolution images, we introduce two networks; one operates at lower-resolution and another only operates on selected patches at the original resolution based on the prediction of the previous network.

The architecture consists of a base network $G_{\text{base}}$ and a refinement network $G_{\text{refine}}$. Given the original image $I$ and the captured background $B$, we first downsample by a factor of $c$ to $I_{c}$ and $B_{c}$. The base network $G_{\text{base}}$ takes $I_{c}$ and $B_{c}$ as input and predicts coarse-grained alpha matte $\alpha_{c}$, foreground residual $F^{R}_{c}$, an error prediction map $E_{c}$, and hidden features $H_{c}$. Then, the refinement network $G_{\text{refine}}$ employs $H_{c}$, $I$, and $B$ to refine $\alpha_{c}$ and $F^{R}_{c}$ only in regions where the predicted error $E_{c}$ is large, and produces alpha $\alpha$ and foreground residual $F^{R}$ at the original resolution. Our model is fully-convolutional and is trained to work on arbitrary sizes and aspect ratios.

\subsection{Base Network}
\label{sec:backbone}
The base network is a fully-convolutional encoder-decoder network inspired by the DeepLabV3 \cite{chen2017rethinking} and DeepLabV3+ \cite{deeplabv3plus2018} architectures, which achieved state-of-the-art performance on semantic segmentation tasks in 2017 and 2018. Our base network consists of three modules: Backbone, ASPP, and Decoder.

We adopt ResNet-50 \cite{he2016resnet} for our encoder backbone, which can be replaced by ResNet-101 and MobileNetV2 \cite{sandler2018mobilenetv2} to trade-off between speed and quality. We adopt the ASPP (Atrous Spatial Pyramid Pooling) module after the backbone following the DeepLabV3 approach. The ASPP module consists of multiple dilated convolution filters with different dilation rates of 3,6 and 9. Our decoder network applies bilinear upsampling at each step, concatenated with the skip connection from the backbone, and followed by a 3$\times$3 convolution, Batch Normalization \cite{ioffe2015batchnorm}, and ReLU activation \cite{nair2010relu} (except the last layer). The decoder network outputs coarse-grained alpha matte $\alpha_{c}$, foreground residual $F^{R}_{c}$, error prediction map $E_{c}$ and a 32-channel hidden features $H_{c}$. The hidden features $H_{c}$ contain global contexts that will be useful for the refinement network.

\subsection{Refinement Network}
\label{sec:refinement}

The goal of the refinement network is to reduce redundant computation and recover high-resolution matting details. While the base network operates on the whole image, the refinement network operates only on patches selected based on the error prediction map $E_{c}$. We perform a two-stage refinement, first at $\frac{1}{2}$ of the original resolution and then at the full resolution.  During inference, we refine $k$ patches, with $k$ either set in advance or set based on a threshold that trades off between quality and computation time. 

Given the coarse error prediction map $E_{c}$ at $\frac{1}{c}$ of the original resolution, we first resample it to $\frac{1}{4}$ of the original resolution $E_{4}$, s.t. each pixel on the map corresponds to a 4$\times$4 patch on the original resolution. We select the top $k$ pixels with the highest predicted error from $E_{4}$ to denote the $k$ 4$\times$4 patch locations that will be refined by our refinement module. The total number of refined pixels at the original resolution is $16k$.

We perform a two-stage refinement process. First, we bilinearly resample the coarse outputs, i.e., alpha matte $\alpha_{c}$, foreground residual $F^{R}_{c}$ and hidden features $H_{c}$, as well as the input image $I$ and background $B$ to $\frac{1}{2}$ of the original resolution and concatenate them as features. Then we crop out 8$\times$8 patches around the error locations selected from $E_{4}$, and pass each through two layers of 3$\times$3 convolution with valid padding, Batch Normalization, and ReLU, which reduce the patch dimension to 4$\times$4. These intermediate features are then upsampled to $8\times8$ again and concatenated with the 8$\times$8 patches extracted from the original-resolution input $I$ and background $B$ at the corresponding location. We then apply an additional two layers of 3$\times$3 convolution with valid padding, Batch Normalization and ReLU (except the last layer) to obtain 4$\times$4 alpha matte and foreground residuals results. Finally, we upsample the coarse alpha matte $\alpha_{c}$ and foreground residual $F^{R}_{c}$ to the original resolution and swap in the respective 4$\times$4 patches that have been refined to obtain the final alpha matte $\alpha$ and foreground residual $F^{R}$. The entire architecture is illustrated in Figure \ref{fig:model_overview}. See appendix for the details of implementation.

\subsection{Training}
\label{sec:training}

All matting datasets provide an alpha matte and a foreground layer, which we compose onto multiple high-resolution backgrounds. We employ multiple data augmentation techniques to avoid overfitting and help the model generalize to challenging real-world situations. We apply affine transformation, horizontal flipping, brightness, hue, and saturation adjustment, blurring, sharpening, and random noise as data augmentation to both the foreground and background layer independently. We also slightly translate the background to simulate misalignment and create artificial shadows to simulate how the presence of a subject can cast shadows in real-life environments (see appendix for more details). We randomly crop the images in every minibatch so that the height and width are each uniformly distributed between 1024 and 2048 to support inference at any resolution and aspect ratio.

To learn $\alpha$ w.r.t. ground-truth $\alpha^{*}$, we use an L1 loss over the whole alpha matte and its (Sobel) gradient:
\vspace{-0.4em}
\begin{equation}
    \mathcal{L}_{\alpha} = || \alpha - \alpha^{*} ||_1 + || \nabla{\alpha} - \nabla{\alpha^{*}} ||_1.
    \vspace{-0.4em}
\end{equation}

We obtain the foreground layer from predicted foreground residual $F^{R}$, using $F = \max(\min(F^{R} + I,1),0)$. We compute L1 loss only on the pixels where $\alpha^{*} > 0$:
\vspace{-0.4em}
\begin{equation}
    \mathcal{L}_{F} = || (\alpha^{*}>0)*(F - F^{*})) ||_1.
    \vspace{-0.4em}
\end{equation}
where that $(\alpha^{*}>0)$ is a Boolean expression.

For refinement region selection, we define the ground truth error map as $E^{*} = | \alpha - \alpha^{*} |$. We then compute mean squared error between the predicted error map and the ground truth error map as the loss:
\vspace{-0.4em}
\begin{equation}
   \mathcal{L}_{E} = || E - E^{*} ||_2.
   \vspace{-0.4em}
\end{equation}
This loss encourages the predicted error map to have larger values where the difference between the predicted alpha and the ground-truth alpha is large. The ground-truth error map changes over iterations during training as the predicted alpha improves. Over time, the error map converges and predicts high error in complex regions, e.g. hair, that would lead to poor composites if simply upsampled.

The base network $(\alpha_{c},F^{R}_{c},E_{c},H_{c}) = G_{\text{base}}(I_{c},B_{c})$ operates at $\frac{1}{c}$ of the original image resolution, and is trained with the following loss function:
\vspace{-0.4em}
\begin{equation}
    \mathcal{L}_{\text{base}} = \mathcal{L}_{\alpha_{c}} + \mathcal{L}_{F_{c}} + \mathcal{L}_{E_{c}}.
    \vspace{-0.4em}
\end{equation}

The refinement network $(\alpha,F^{R}) = G_{\text{refine}}(\alpha_{c},F^{R}_{c},E_{c},H_{c},I,B)$ is trained using:
\vspace{-0.4em}
\begin{equation}
    \mathcal{L}_{\text{refine}} = \mathcal{L}_{\alpha} + \mathcal{L}_{F} .
    \vspace{-0.4em}
\end{equation}

We initialize our model's backbone and ASPP module with DeepLabV3 weights pre-trained for semantic segmentation on ImageNet and Pascal VOC datasets. We first train the base network till convergence and then add the refinement module and train it jointly. We use Adam optimizer and $c=4$, $k=5,000$ during all the training. For training only the base network, we use batch size 8 and learning rate [1e-4, 5e-4, 5e-4] for backbone, ASPP, and decoder. When training jointly, we use batch size 4 and learning rate [5e-5, 5e-5, 1e-4, 3e-4] for backbone, ASPP, decoder, and refinement module respectively.

We train our model on multiple datasets in the following order. First, we train only the base network $G_{\text{base}}$ and then the entire model $G_{\text{base}}$ and $G_{\text{refine}}$ jointly on VideoMatte240K, which makes the model robust to a variety of subjects and poses. Next, we train our model jointly on PhotoMatte13K to improve the high-resolution details. Finally, we train our model jointly on Distinctions-646. The dataset has only 362 unique training samples, but it is of the highest quality and contains human-annotated foregrounds that are very helpful for improving the foreground quality produced by our model. We omit training on the AIM dataset as a possible 4th stage and only use it for testing because it causes a degradation in quality as shown in our ablation study in Section \ref{sec:abla}.
\section{Experimental Evaluation}
\label{sec:exp-sec}

\begin{table}[t]
  \begin{center}
    \setlength\tabcolsep{2 pt}
    \begin{tabular}{ll|rrrr|r}
      \toprule
      \multicolumn{2}{c|}{} & \multicolumn{4}{c|}{Alpha} & \multicolumn{1}{c}{FG} \\
      Dataset & Method & SAD & MSE & Grad & Conn & MSE \\
      \midrule
      \multirow{5.5}{*}{AIM}
        & DIM$^{\text{\textdagger}}$ & 37.94 & 80.67 & 32935 & 37861 & - \\
        & FBA$^{\text{\textdagger}}$ & \textbf{9.68} & \textbf{6.38} & \textbf{4265} & \textbf{7521} & \textbf{1.94} \\
        \cmidrule{2-7}
        & BGM & 16.07 & 21.00 & 15371 & 14123 & 47.98 \\
        & BGM$_a$ & 19.28 & 29.31 & 19877 & 18083 & 42.84 \\
        & Ours & \textbf{12.86} & \textbf{12.01} & \textbf{8426} & \textbf{11116} & \textbf{5.31}\\
      \midrule
      \multirow{5.5}{*}{Distinctions}
        & DIM$^{\text{\textdagger}}$ & 43.70 & 86.22 & 49739 & 43914 & -\\
        & FBA$^{\text{\textdagger}}$ & \textbf{11.03} & \textbf{8.32} & \textbf{6894} & \textbf{9892} & \textbf{12.51}\\
        \cmidrule{2-7}
        & BGM & 19.21 & 25.89 & 30443 & 18191 & 36.13\\
        & BGM$_a$ & 16.02 & 20.18 & 24845 & 14900 & 43.00\\
        & Ours & \textbf{9.19} & \textbf{7.08} & \textbf{6345} & \textbf{7216} & \textbf{6.10} \\
      \midrule
      \multirow{5.5}{*}{PhotoMatte85}
        & DIM$^{\text{\textdagger}}$ & 32.26 & 45.40 & 44658 & 30876 & -\\
        & FBA$^{\text{\textdagger}}$ & \textbf{7.37} & \textbf{4.79} & \textbf{7323} & \textbf{5206} & \textbf{7.03}\\
        \cmidrule{2-7}
        & BGM & 17.32 & 21.21 & 27454 & 15397 & 14.25\\
        & BGM$_a$ & 14.45 & 19.24 & 23314 & 13091 & 16.80\\
        & Ours & \textbf{8.65} & \textbf{9.57} & \textbf{8736} & \textbf{6637} & \textbf{13.82}\\
      \bottomrule
    \end{tabular}
    \vspace{-0.5em}
    \caption{Quantitative evaluation on different datasets. $^{\text{\textdagger}}$ indicates methods that require a manual trimap.}
    \label{tab:quantitative-evaluation}
  \end{center}
      \vspace{-2.5em}
\end{table}

\begin{figure*}[h!]
    \centering
    \includegraphics[width=0.95\textwidth]{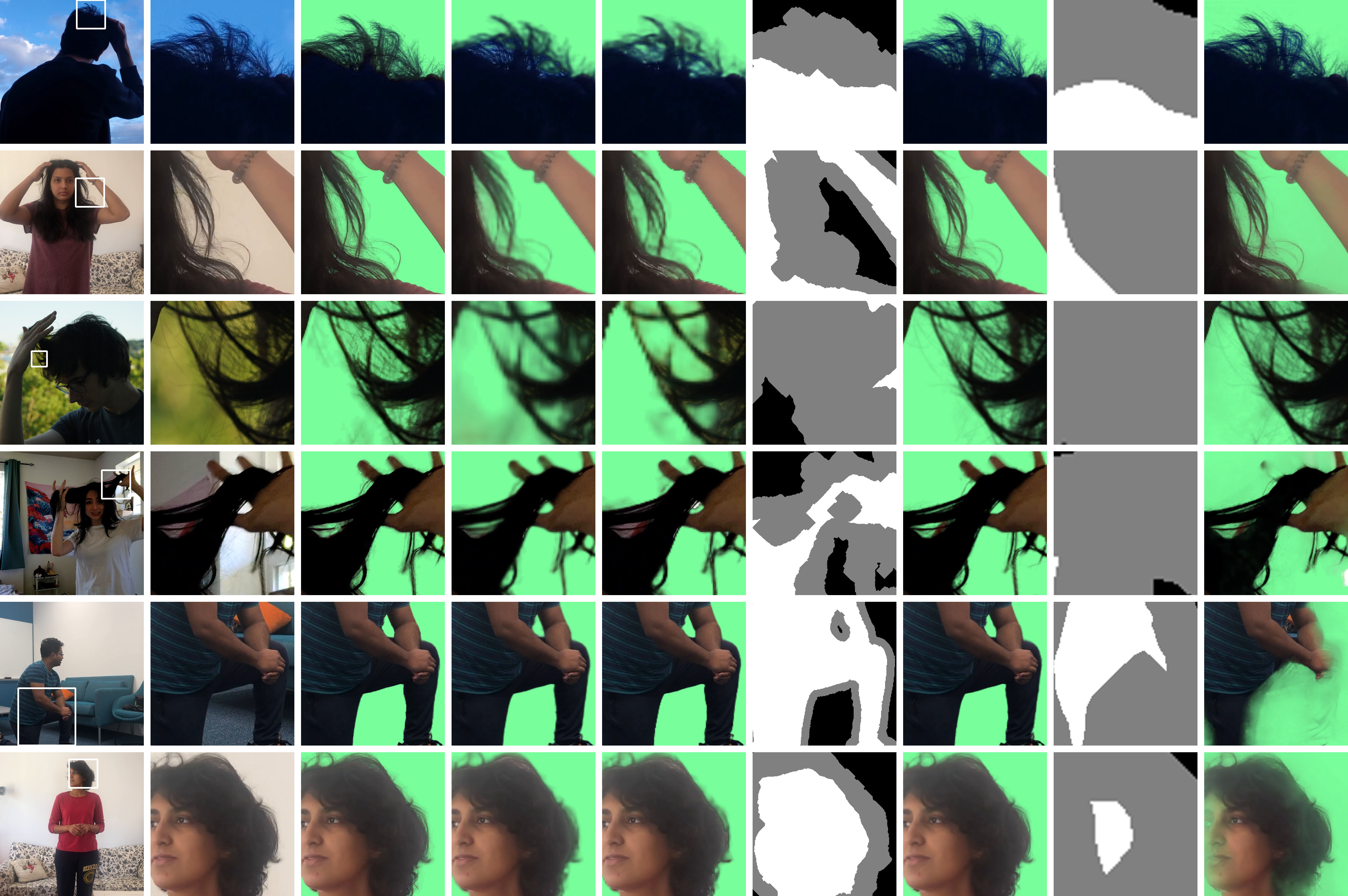}
    \begin{small}
    \begin{tabularx}{.95\textwidth}{XX|XXX|XX|XX}
        & Input & Ours & BGM & BGM$_a$ & Trimap & FBA & Trimap & FBA$_{\text{auto}}$ \\
        & & \multicolumn{3}{l|}{\footnotesize{Background-based methods}} & \multicolumn{2}{l|}{\footnotesize{Manual trimap}} & \multicolumn{2}{l}{\footnotesize{Segmentation-morph trimap}}
    \end{tabularx}
    \end{small}
    \vspace{-0.5em}
    \caption{Qualitative comparison on real images. We produce superior results at high-resolution with minimal user input. While FBA is competitive, it fails in a fully automatic application where the segmentation-based trimap is faulty.}
    \label{fig:exp-qualitative-evaluation}
    \vspace{-1.5em}
\end{figure*}

We compare our approach to two trimap-based methods, Deep Image Matting (DIM) \cite{xu2017deep} and FBA Matting (FBA) \cite{fbaMatting20}, and one background-based method, Background Matting (BGM) \cite{sengupta2020background}. The input resolution to DIM was fixed at 320$\times$320 by the implementation, while we set the FBA input resolution to approximately HD due to memory limits. We additionally train the BGM model on our datasets and denote it as BGM$_a$ (BGM adapted).

Our evaluation uses $c=4$, $k=20,000$ for photos, $c=4$, $k=5,000$ for HD videos, and $c=8$, $k=20,000$ for 4K videos, where $c$ is the downsampling factor for the base network and $k$ is the number of patches that get refined.

\subsection{Evaluation on composition datasets}
\label{sec:quants}
We construct test benchmarks by separately compositing test samples from AIM, Distinctions, and PhotoMatte85 datasets onto 5 background images per sample. We apply minor background misalignment, color adjustment, and noise to simulate flawed background capture. We generate trimaps from ground-truth alpha using thresholding and morphological operations. We evaluate matte outputs using metrics from \cite{rhemann2009perceptually}: MSE (mean squared error) for alpha and foreground, SAD (sum of absolute difference), Grad (spatial-gradient metric), and Conn (connectivity) for alpha only. 
All MSE values are scaled by 10$^3$ and all metrics are only computed over the unknown region of trimaps generated as described above. Foreground MSE is additionally only measured where the grouth-truth alpha $\alpha^* > 0$.

Table \ref{tab:quantitative-evaluation} shows that our approach outperforms the existing background-based BGM method across all datasets. Our approach is slightly worse than the state-of-the-art trimap-based FBA method, which requires carefully annotated manual trimaps and is much slower than our approach, as shown later in the performance comparison.

\subsection{Evaluation on captured data}
\label{sec:quals}

Although quantitative evaluation on the above-mentioned datasets serves the purpose of quantifying the performance of different algorithms, it is important to evaluate how these methods perform on unconstrained real data. To evaluate on real data, we capture a number of photos and videos containing subjects in varying poses and surroundings. The videos are captured on a tripod with consumer smartphones (Samsung S10+ and iPhone X) and a professional camera (Sony $\alpha$7s II), in both HD and 4K resolution. The photos are captured in 4000$\times$6000 resolution. We also use some HD videos presented in the BGM paper that are made public to compare with our method.

For fair comparison in the real-time scenario, where manual trimaps cannot be crafted, we construct trimaps by morphing segmentation result from DeepLabV3+, as suggested in \cite{sengupta2020background}. We show results on both trimaps, denoting FBA using this fully automatic trimap as FBA$_{\text{auto}}$.

Figure \ref{fig:exp-qualitative-evaluation} shows our method produces much sharper and more detailed results around hair and edges compared to other methods. Since our refinement operates at the native resolution, the quality is far superior relative to BGM, which resizes the images and only processes them at 512$\times$512 resolution. FBA, with manual trimap, produces excellent results around hair details, however cannot be evaluated at resolutions above around HD on standard GPUs. When FBA is applied on automatic trimaps generated with segmentation, it often shows large artifacts, mainly due to faulty segmentation.

We extract 34 frames from both the test videos shared by the BGM paper and our captured videos and photos to create a user study. 40 participants were presented with an interactive interface showing each input image as well as the mattes produced by BGM and our approach, in random order. They were encouraged to zoom in on details and asked to rate one of the mattes as "much better", "slightly better", or "similar". The results, shown in Table \ref{tab:userstudy}, demonstrate significant qualitative improvement over BGM. 59\% of the time participants perceive our algorithm to be better, compared to 23\% for BGM. For sharp samples in 4K and larger, our method is preferred 75\% of the time to BGM's 15\%.

\begin{table}[h!]
  \vspace{-0.5em}
  \setlength\tabcolsep{3.5 pt}
  \begin{center}
    \begin{tabular}{lccccc}
      \toprule
      & Much worse & Worse & Similar & Better & Much better \\
      \midrule
      All & 6\% & 17\% & 18\% & 32\% & 27\% \\
      4K+ & 5\% & 10\% & 10\% & 34\% & 41\% \\
      \bottomrule
    \end{tabular}
    \vspace{-0.5em}
    \caption{User study results: Ours vs BGM}
    \label{tab:userstudy}
  \end{center}
  \vspace{-2em}
\end{table}

\subsection{Performance comparison}
\label{sec:runtime}

Table \ref{tab:exp-model-performance} and \ref{tab:exp-model-size} show that our method is smaller and much faster than BGM. Our method contains only 55.7\% of the parameters compared to BGM. Our method can achieve HD 60fps and 4K 30fps at batch size 1 on an Nvidia RTX 2080 TI GPU, considered to be real-time for many applications. It is a significant speed-up compared to BGM which can only handle 512$\times$512 resolution at 7.8fps. The performance can be further improved by switching to MobileNetV2 backbone, which achieves 4K 45fps and HD 100fps. More performance results, such as adjusting the refinement selection parameter $k$ and using a larger batch size, are included in the ablation studies and in the appendix.

\begin{table}[!h]
  \begin{center}
  \setlength\tabcolsep{4.5 pt}
    \begin{tabularx}{.45\textwidth}{llcrr}
      \toprule
      Method & Backbone & Resolution & FPS & GMac \\
      \midrule
      FBA & & HD & 3.3 & 54.3 \\
      FBA$_{\text{auto}}$ & & HD & 2.9 & 137.6 \\
      \midrule
      BGM & & 512$^2$ & 7.8 & 473.8 \\
      \midrule
      \multirow{3}{*}{Ours}
       & ResNet-50* & HD & 60.0 & 34.3  \\
       & ResNet-101 & HD & 42.5 & 44.0  \\
       & MobileNetV2 & HD & 100.6 & 9.9 \\
      \midrule
      \multirow{3}{*}{Ours}
       & ResNet-50* & 4K & 33.2 & 41.5  \\
       & ResNet-101 & 4K & 29.8 & 51.2 \\
       & MobileNetV2 & 4K & 45.4 & 17.0  \\
      \bottomrule
    \end{tabularx}
    \vspace{-0.5em}
    \caption{Speed measured on Nvidia RTX 2080 TI as PyTorch model pass-through without data transferring at FP32 precision and with batch size 1. GMac does not account for interpolation and cropping operations. For the ease of measurement, BGM and FBA$_{\text{auto}}$ use adapted PyTorch DeepLabV3+ implementation with ResNet101 backbone as segmentation.}
    \label{tab:exp-model-performance}
  \end{center}
  \vspace{-2.0em}
\end{table}

\begin{table}[!h]
  \begin{center}
  \setlength\tabcolsep{6 pt}
    \begin{tabularx}{.45\textwidth}{llrr}
      \toprule
      Method & Backbone & Parameters & Size \\
      \midrule
      FBA & & 34,693,031 & 138.80 MB \\
      FBA$_{\text{auto}}$ & & 89,398,348 & 347.48 MB \\
      \midrule
      BGM & & 72,231,209 & 275.53 MB \\
      \midrule
      \multirow{3}{*}{Ours}
       & ResNet-50* & 40,247,703 & 153.53 MB \\
       & ResNet-101 & 59,239,831 & 225.98 MB \\
       & MobileNetV2 & 5,006,839 & 19.10 MB \\
      \bottomrule
    \end{tabularx}
    \vspace{-0.5em}
    \caption{Model size comparison. BGM and FBA$_{\text{auto}}$ use DeepLabV3+ with Xception backbone for segmentation.}
    \label{tab:exp-model-size}
  \end{center}
  \vspace{-2em}
\end{table}

\begin{figure}[h!]
    \centering
    \newcolumntype{A}{>{\hsize=0.202\textwidth}X}
    \newcolumntype{B}{>{\hsize=0.086\textwidth}X}
    \begin{footnotesize}
        \begin{tabularx}{.45\textwidth}{ABB}
          Natural capture & BGM & Ours
        \end{tabularx}
    \end{footnotesize}
    \includegraphics[width=0.45\textwidth]{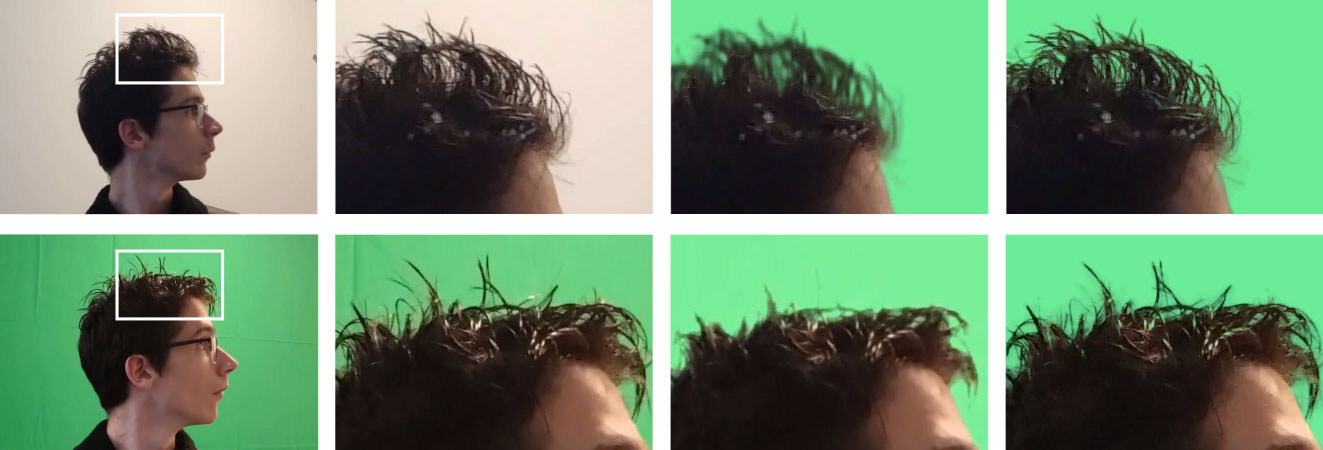}
    \begin{footnotesize}
        \begin{tabularx}{.45\textwidth}{ABB}
          Green screen capture & DaVinci & Ours
        \end{tabularx}
    \end{footnotesize}
    \vspace{-0.5em}
    \caption{We produce better results than a chroma-keying software, when an amateur green-screen setup is used.}
    \label{fig:exp-green-screen}
    \vspace{-1.5em}
\end{figure}

\subsection{Practical use}
\textbf{Zoom implementation} We build a Zoom plugin that intercepts the webcam input, collects one no-person (background) shot, then performs real-time video matting and compositing, streaming the result back into the Zoom call. We test with a 720p webcam in Linux. The upgrade elicits praise in real meetings, demonstrating its practicality in a real-world setting.

\textbf{Comparison to green-screen} Chroma keying with a green screen is the most popular method for creating high-quality mattes. However, it requires even lighting across the screen and background-subject separation to avoid cast shadows. In Figure \ref{fig:exp-green-screen}, we compare our method against chroma-keying under the same lighting with an amateur green-screen setup. We find that in the unevenly lit setting, our method outperforms approaches designed for the green screen.

\section{Ablation Studies}
\label{sec:abla}
\textbf{Role of our datasets} We train on multiple datasets, each of which brings unique characteristics that help our network produce high-quality results at high-resolution. Table \ref{tab:abla-dataset} shows the metrics of our method by adding or removing a dataset from our training pipeline. We find adding the AIM dataset as a possible 4th stage worsens the metrics even on the AIM test set itself. We believe it is because samples in the AIM dataset are lower in resolution and quality compared to Distinctions and the small number of samples may have caused overfitting. Removal of VideoMatte240K, PhotoMatte13K, and Distinctions datasets from the training pipeline all result in worse metrics, proving that those datasets are helpful in improving the model's quality.
\begin{table}[t]
  \begin{center}
    \setlength\tabcolsep{2.5 pt}
    \begin{tabularx}{.45\textwidth}{l|rrrr|r}
      \toprule
      \multicolumn{1}{c|}{} & \multicolumn{4}{c|}{Alpha} & \multicolumn{1}{c}{FG} \\
      Method & SAD & MSE & Grad & Conn & MSE \\
      \midrule
      Ours* & \textbf{12.86} & \textbf{12.01} & \textbf{8426} & \textbf{11116} & \textbf{5.31} \\
      + AIM & 14.19 & 14.70 & 9629 & 12648 & 6.34\\
      - PhotoMatte13K & 14.05 & 14.10 & 10102 & 12749 & 6.53 \\
      - VideoMatte240K & 15.17 & 17.31 & 11907 & 13827 & 7.04 \\
      - Distinctions & 15.95 & 19.51 & 11911 & 14909 & 14.36 \\
      \midrule
      BGM & 16.07 & 21.00 & 15371 & 14123 & 42.84\\
      \bottomrule
    \end{tabularx}
    \vspace{-0.5em}
    \caption{Effect of removing or appending a dataset in the training pipeline, evaluated on the AIM test set.}
    \label{tab:abla-dataset}
  \end{center}
  \vspace{-1em}
\end{table}

\textbf{Role of the base network} We experiment with replacing ResNet-50 with ResNet-101 and MobileNetV2 as our encoder backbone in the base network. The metrics in Table \ref{tab:abla-base-and-refine} show that ResNet-101 has slight improvements over ResNet-50 on some metrics while doing worse on others. This indicates that ResNet-50 is often sufficient for obtaining the best quality. MobileNetV2 on the other hand is worse than ResNet-50 on all metrics, but it is significantly faster and smaller than ResNet-50 as shown in Tables \ref{tab:exp-model-performance} and \ref{tab:exp-model-size}, and still obtains better metrics than BGM.

\begin{table}[t]
    \centering
    \setlength\tabcolsep{1.5 pt}
    \begin{tabularx}{.45\textwidth}{lc|rrrr|r}
         \toprule
         Base & \multicolumn{1}{c|}{Refine} & \multicolumn{4}{c|}{Alpha} & \multicolumn{1}{c}{FG} \\
         Backbone & Kernel & SAD & MSE & Grad & Conn & MSE  \\
         \midrule
         BGM$_a$ & & 16.02 & 20.18 & 24845 & 14900 & 43.00\\
         \midrule
         MobileNetV2 & 3$\times$3 & 10.53 & 9.62 & 7904 & 8808 & 8.19 \\
         ResNet-50* & 3$\times$3 & \textbf{9.19} & 7.08 & 6345 & 7216 & \textbf{6.10} \\
         ResNet-101 & 3$\times$3 & 9.30 & \textbf{6.82} & \textbf{6191} & \textbf{7128} & 7.68 \\
         \midrule
         ResNet-50 & 1$\times$1 & 9.36 & 8.06 & 7319 & 7395 & 6.92 \\ 
         \bottomrule
    \end{tabularx}
    \vspace{-0.5em}
    \caption{Comparison of backbones and refinement kernels on the Distinctions test set}
    \label{tab:abla-base-and-refine}
    \vspace{-1.5em}
\end{table}

\begin{figure}[h!]
    \centering
    \includegraphics[width=0.45\textwidth]{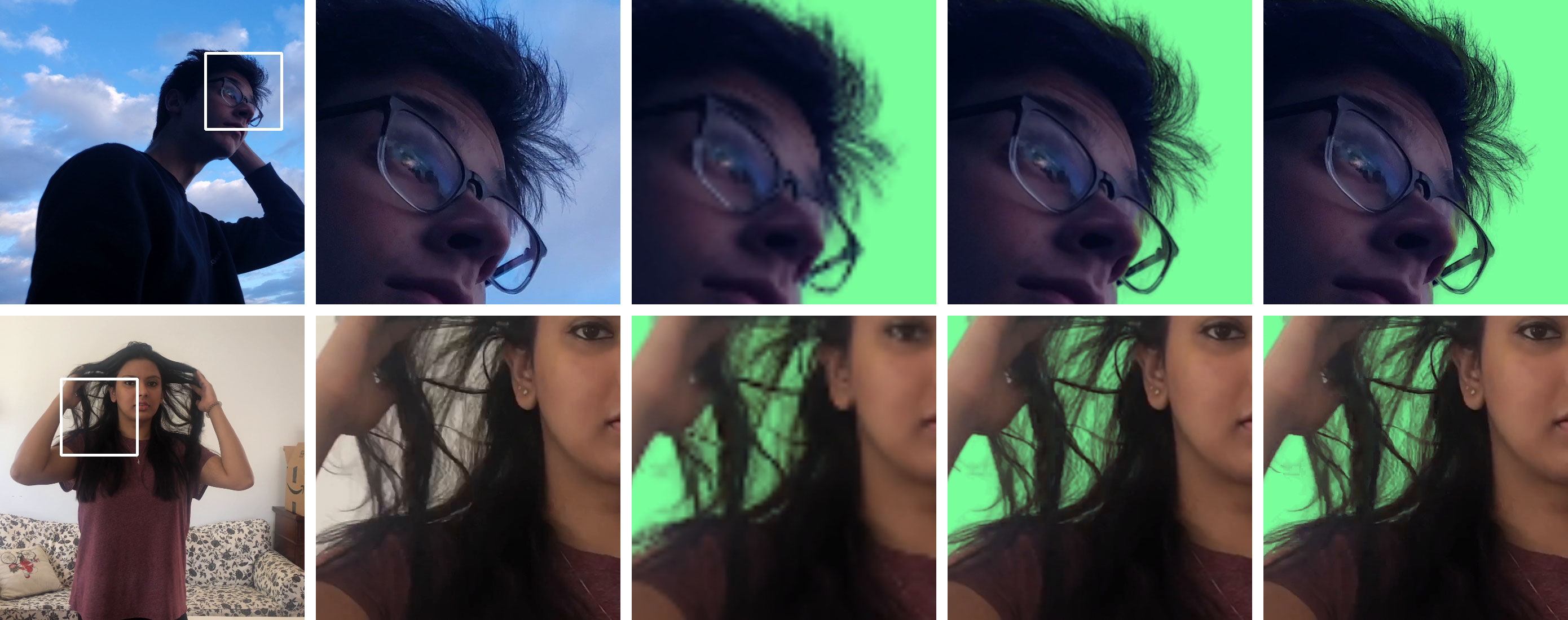}
    \begin{footnotesize}
        \begin{tabularx}{.45\textwidth}{XXXXX}
             & Input & 480$\times$270 & HD & 4K
        \end{tabularx}
    \end{footnotesize}
    \vspace{-0.8em}
    \caption{Effect of refinement, from coarse to HD and 4K.}
    \label{fig:refinement_to_4k}
    \vspace{-0.5em}
\end{figure}

\begin{figure}[h!]
    \centering
    \includegraphics[width=0.45\textwidth]{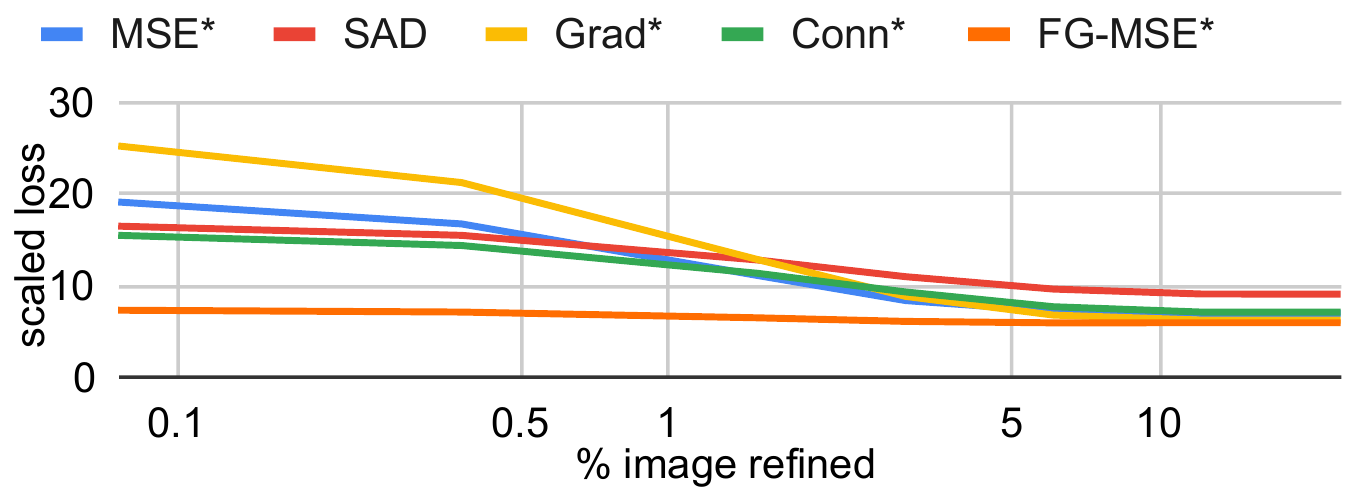}
    \vspace{-0.5em}
    \caption{Metrics on the Distinctions test set over percentage of image area refined. Grad and Conn are scaled by 10$^{-3}$.}
    \label{fig:sample-pixels}
\end{figure}

\textbf{Role of the refinement network} Our refinement network improves detail sharpness over the coarse results in Figure \ref{fig:refinement_to_4k}, and is effective even in 4K resolution. Figure \ref{fig:sample-pixels} shows the effects of increasing and decreasing the refinement area. Most improvement can be achieved by refining over only 5\% to 10\% of the image resolution. Table \ref{tab:sample-pixels} shows that refining only the selected patches provides significant speedup compared to refining the full image.

\vspace{-0.5em}

\begin{table}[h!]
    \centering
    \setlength\tabcolsep{11 pt}
    \begin{tabularx}{.45\textwidth}{l|cccc}
        \toprule
        $k$ & 2,500 & 5,000* & 7,500 & Full  \\
        \midrule
        FPS & 62.9 & 60.0 & 55.7 & 42.8 \\
        \bottomrule
    \end{tabularx}
    \vspace{-0.5em}
    \caption{Performance with different $k$ values. Measured on our method with ResNet-50 backbone at HD.}
    \label{tab:sample-pixels}
\end{table}

\textbf{Patch-based refinement vs. Point-based refinement}  Our refinement module uses a stack of 3$\times$3 convolution kernels, creating a 13$\times$13 receptive field for every output pixel. An alternative is to refine only on points using 1$\times$1 convolution kernels, which would result in a 2$\times$2 receptive field with our method. Table \ref{tab:abla-base-and-refine} shows that the $3\times3$ kernel can achieve better metrics than point-based kernels, due to a larger receptive field.

\begin{figure}[t]
    \centering
    \includegraphics[width=0.45\textwidth]{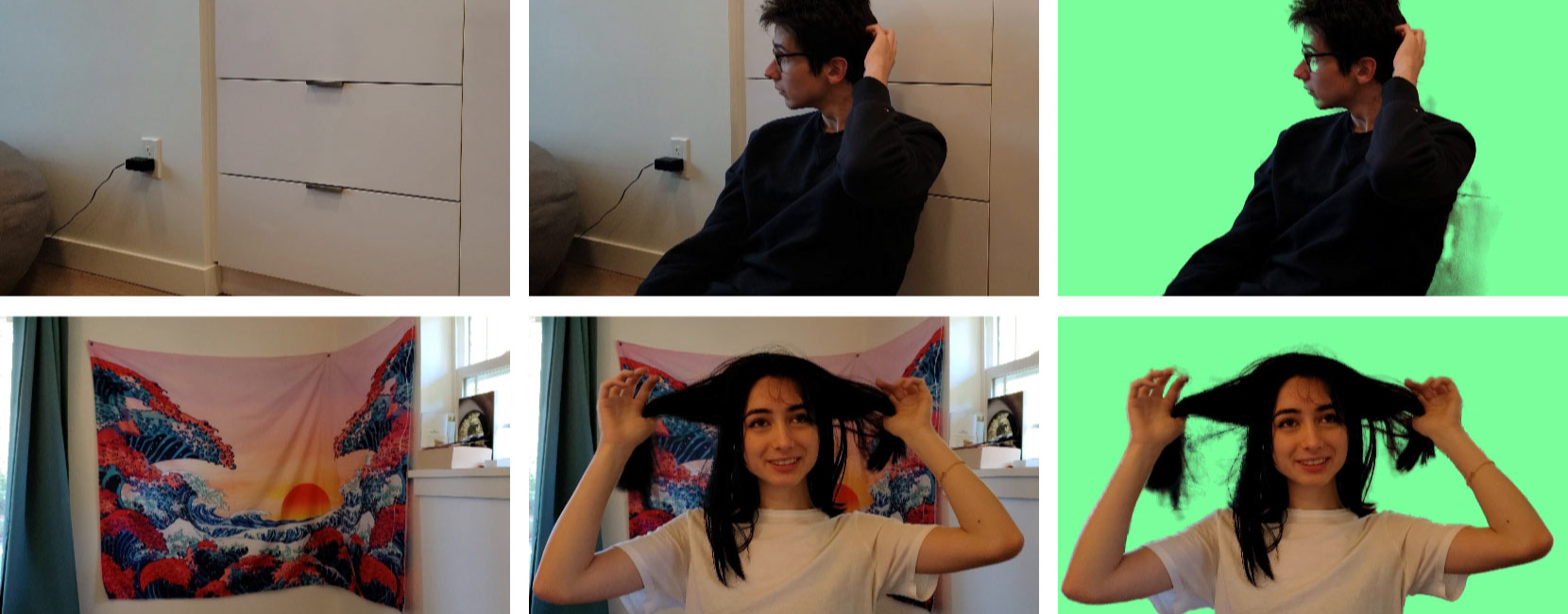}
    \caption{Failure cases. Our method fails when the subject casts a substantial shadow on, or matches color with, the background (top) and when the background is highly textured (bottom).}
    \label{fig:abla-failure}
    \vspace{-1.0em}
\end{figure}
\textbf{Limitations} Our method can be used on handheld input by applying homography alignment to the background on every frame, but it is limited to small motion. Other common limitations are indicated in Figure \ref{fig:abla-failure}. We recommend using our method with a simple-textured background, fixed exposure/focus/WB setting, and a tripod for the best result.

\section{Conclusion}
\label{sec:discussion}

We have proposed a real-time, high-resolution background replacement technique that operates at 4K 30fps and HD 60fps. Our method only requires an input image and an pre-captured background image, which is easy to obtain in many applications. Our proposed architecture efficiently refines only the error-prone regions at high-resolution, which reduces redundant computation and makes real-time high-resolution matting possible. We introduce two new large-scale matting datasets that help our method generalize to real-life scenarios. Our experiment shows our method sets new state-of-the-art performance on background matting. We demonstrate the practicality of our method by streaming our results to Zoom and achieve a much more realistic virtual conference call.

\textbf{Ethics} Our primary goal is to enable creative applications and give users more privacy options through background replacement in video calls.  However, we recognize that image editing can also be used for negative purposes, which can be mitigated through watermarking and other security techniques in commercial applications of this work.

\clearpage
\newpage

{\small
\bibliographystyle{ieee_fullname}
\bibliography{ref}

\begin{thebibliography}{10}\itemsep=-1pt

\bibitem{aksoy2018semantic}
Ya{\u{g}}iz Aksoy, Tae-Hyun Oh, Sylvain Paris, Marc Pollefeys, and Wojciech
  Matusik.
\newblock Semantic soft segmentation.
\newblock {\em ACM Transactions on Graphics (TOG)}, 37(4):72, 2018.

\bibitem{aksoy2017designing}
Yagiz Aksoy, Tunc Ozan~Aydin, and Marc Pollefeys.
\newblock Designing effective inter-pixel information flow for natural image
  matting.
\newblock In {\em Proceedings of the IEEE Conference on Computer Vision and
  Pattern Recognition}, pages 29--37, 2017.

\bibitem{cai2019disentangled}
Shaofan Cai, Xiaoshuai Zhang, Haoqiang Fan, Haibin Huang, Jiangyu Liu, Jiaming
  Liu, Jiaying Liu, Jue Wang, and Jian Sun.
\newblock Disentangled image matting.
\newblock {\em International Conference on Computer Vision (ICCV)}, 2019.

\bibitem{chen2017rethinking}
Liang-Chieh Chen, George Papandreou, Florian Schroff, and Hartwig Adam.
\newblock Rethinking atrous convolution for semantic image segmentation.
\newblock {\em arXiv preprint arXiv:1706.05587}, 2017.

\bibitem{deeplabv3plus2018}
Liang-Chieh Chen, Yukun Zhu, George Papandreou, Florian Schroff, and Hartwig
  Adam.
\newblock Encoder-decoder with atrous separable convolution for semantic image
  segmentation.
\newblock In {\em ECCV}, 2018.

\bibitem{chen2013knn}
Qifeng Chen, Dingzeyu Li, and Chi-Keung Tang.
\newblock Knn matting.
\newblock {\em IEEE transactions on pattern analysis and machine intelligence},
  35(9):2175--2188, 2013.

\bibitem{chuang2001bayesian}
Yung-Yu Chuang, Brian Curless, David~H Salesin, and Richard Szeliski.
\newblock A bayesian approach to digital matting.
\newblock In {\em CVPR (2)}, pages 264--271, 2001.

\bibitem{fbaMatting20}
Marco Forte and François Pitié.
\newblock F,b, alpha matting.
\newblock {\em arXiv preprint arXiv:2003.07711}, 2020.

\bibitem{gastal2010shared}
Eduardo~SL Gastal and Manuel~M Oliveira.
\newblock Shared sampling for real-time alpha matting.
\newblock In {\em Computer Graphics Forum}, volume~29, pages 575--584. Wiley
  Online Library, 2010.

\bibitem{gong2009near}
Minglun Gong and Yee-Hong Yang.
\newblock Near-real-time image matting with known background.
\newblock In {\em 2009 Canadian Conference on Computer and Robot Vision}, pages
  81--87. IEEE, 2009.

\bibitem{he2017mask}
Kaiming He, Georgia Gkioxari, Piotr Doll{\'a}r, and Ross Girshick.
\newblock Mask r-cnn.
\newblock In {\em Proceedings of the IEEE international conference on computer
  vision}, pages 2961--2969, 2017.

\bibitem{he2016resnet}
Kaiming He, X. Zhang, Shaoqing Ren, and Jian Sun.
\newblock Deep residual learning for image recognition.
\newblock {\em 2016 IEEE Conference on Computer Vision and Pattern Recognition
  (CVPR)}, pages 770--778, 2016.

\bibitem{hou2019context}
Qiqi Hou and Feng Liu.
\newblock Context-aware image matting for simultaneous foreground and alpha
  estimation.
\newblock {\em International Conference on Computer Vision (ICCV)}, 2019.

\bibitem{ioffe2015batchnorm}
S. Ioffe and Christian Szegedy.
\newblock Batch normalization: Accelerating deep network training by reducing
  internal covariate shift.
\newblock {\em ArXiv}, abs/1502.03167, 2015.

\bibitem{kirillov2020pointrend}
Alexander Kirillov, Yuxin Wu, Kaiming He, and Ross Girshick.
\newblock Pointrend: Image segmentation as rendering.
\newblock In {\em Proceedings of the IEEE/CVF Conference on Computer Vision and
  Pattern Recognition}, pages 9799--9808, 2020.

\bibitem{levin2007closed}
Anat Levin, Dani Lischinski, and Yair Weiss.
\newblock A closed-form solution to natural image matting.
\newblock {\em IEEE transactions on pattern analysis and machine intelligence},
  30(2):228--242, 2007.

\bibitem{levin2008spectral}
Anat Levin, Alex Rav-Acha, and Dani Lischinski.
\newblock Spectral matting.
\newblock {\em IEEE transactions on pattern analysis and machine intelligence},
  30(10):1699--1712, 2008.

\bibitem{li2020hierarchical}
Yaoyi Li, Qingyao Xu, and Hongtao Lu.
\newblock Hierarchical opacity propagation for image matting.
\newblock {\em arXiv preprint arXiv:2004.03249}, 2020.

\bibitem{liang2018look}
Xiaodan Liang, Ke Gong, Xiaohui Shen, and Liang Lin.
\newblock Look into person: Joint body parsing \& pose estimation network and a
  new benchmark.
\newblock {\em IEEE transactions on pattern analysis and machine intelligence},
  41(4):871--885, 2018.

\bibitem{liu2020boosting}
Jinlin Liu, Yuan Yao, Wendi Hou, Miaomiao Cui, Xuansong Xie, Changshui Zhang,
  and Xian-sheng Hua.
\newblock Boosting semantic human matting with coarse annotations.
\newblock In {\em Proceedings of the IEEE/CVF Conference on Computer Vision and
  Pattern Recognition}, pages 8563--8572, 2020.

\bibitem{lu2019indices}
Hao Lu, Yutong Dai, Chunhua Shen, and Songcen Xu.
\newblock Indices matter: Learning to index for deep image matting.
\newblock {\em International Conference on Computer Vision (ICCV)}, 2019.

\bibitem{nair2010relu}
V. Nair and Geoffrey~E. Hinton.
\newblock Rectified linear units improve restricted boltzmann machines.
\newblock In {\em ICML}, 2010.

\bibitem{pytorch}
Adam Paszke, S. Gross, Francisco Massa, A. Lerer, J. Bradbury, G. Chanan, T.
  Killeen, Z. Lin, N. Gimelshein, L. Antiga, Alban Desmaison, Andreas K{\"o}pf,
  E. Yang, Zach DeVito, Martin Raison, Alykhan Tejani, Sasank Chilamkurthy, B.
  Steiner, Lu Fang, Junjie Bai, and Soumith Chintala.
\newblock Pytorch: An imperative style, high-performance deep learning library.
\newblock {\em ArXiv}, abs/1912.01703, 2019.

\bibitem{qian1999video}
Richard~J Qian and M~Ibrahim Sezan.
\newblock Video background replacement without a blue screen.
\newblock In {\em Proceedings 1999 International Conference on Image Processing
  (Cat. 99CH36348)}, volume~4, pages 143--146. IEEE, 1999.

\bibitem{qiao2020attention}
Yu Qiao, Yuhao Liu, Xin Yang, Dongsheng Zhou, Mingliang Xu, Qiang Zhang, and
  Xiaopeng Wei.
\newblock Attention-guided hierarchical structure aggregation for image
  matting.
\newblock In {\em Proceedings of the IEEE/CVF Conference on Computer Vision and
  Pattern Recognition}, pages 13676--13685, 2020.

\bibitem{rhemann2009perceptually}
Christoph Rhemann, Carsten Rother, Jue Wang, Margrit Gelautz, Pushmeet Kohli,
  and Pamela Rott.
\newblock A perceptually motivated online benchmark for image matting.
\newblock In {\em 2009 IEEE Conference on Computer Vision and Pattern
  Recognition}, pages 1826--1833. IEEE, 2009.

\bibitem{sandler2018mobilenetv2}
Mark Sandler, Andrew Howard, Menglong Zhu, Andrey Zhmoginov, and Liang-Chieh
  Chen.
\newblock Mobilenetv2: Inverted residuals and linear bottlenecks.
\newblock In {\em Proceedings of the IEEE conference on computer vision and
  pattern recognition}, pages 4510--4520, 2018.

\bibitem{sengupta2020background}
Soumyadip Sengupta, Vivek Jayaram, Brian Curless, Steven~M Seitz, and Ira
  Kemelmacher-Shlizerman.
\newblock Background matting: The world is your green screen.
\newblock In {\em Proceedings of the IEEE/CVF Conference on Computer Vision and
  Pattern Recognition}, pages 2291--2300, 2020.

\bibitem{shen2016deep}
Xiaoyong Shen, Xin Tao, Hongyun Gao, Chao Zhou, and Jiaya Jia.
\newblock Deep automatic portrait matting.
\newblock In {\em European Conference on Computer Vision}, pages 92--107.
  Springer, 2016.

\bibitem{sun2004poisson}
Jian Sun, Jiaya Jia, Chi-Keung Tang, and Heung-Yeung Shum.
\newblock Poisson matting.
\newblock In {\em ACM Transactions on Graphics (ToG)}, volume~23, pages
  315--321. ACM, 2004.

\bibitem{Tang_2019_CVPR}
Jingwei Tang, Yagiz Aksoy, Cengiz Oztireli, Markus Gross, and Tunc~Ozan Aydin.
\newblock Learning-based sampling for natural image matting.
\newblock In {\em The IEEE Conference on Computer Vision and Pattern
  Recognition (CVPR)}, June 2019.

\bibitem{wang2008image}
Jue Wang, Michael~F Cohen, et~al.
\newblock Image and video matting: a survey.
\newblock {\em Foundations and Trends{\textregistered} in Computer Graphics and
  Vision}, 3(2):97--175, 2008.

\bibitem{xu2017deep}
Ning Xu, Brian Price, Scott Cohen, and Thomas Huang.
\newblock Deep image matting.
\newblock In {\em Proceedings of the IEEE Conference on Computer Vision and
  Pattern Recognition}, pages 2970--2979, 2017.

\bibitem{zhang2019pose2seg}
Song-Hai Zhang, Ruilong Li, Xin Dong, Paul Rosin, Zixi Cai, Xi Han, Dingcheng
  Yang, Haozhi Huang, and Shi-Min Hu.
\newblock Pose2seg: Detection free human instance segmentation.
\newblock In {\em Proceedings of the IEEE conference on computer vision and
  pattern recognition}, pages 889--898, 2019.

\bibitem{zhang2019late}
Yunke Zhang, Lixue Gong, Lubin Fan, Peiran Ren, Qixing Huang, Hujun Bao, and
  Weiwei Xu.
\newblock A late fusion cnn for digital matting.
\newblock In {\em Proceedings of the IEEE Conference on Computer Vision and
  Pattern Recognition}, pages 7469--7478, 2019.

\bibitem{zhu2017fast}
Bingke Zhu, Yingying Chen, Jinqiao Wang, Si Liu, Bo Zhang, and Ming Tang.
\newblock Fast deep matting for portrait animation on mobile phone.
\newblock In {\em Proceedings of the 25th ACM international conference on
  Multimedia}, pages 297--305. ACM, 2017.

\end{thebibliography}
}

\clearpage
\renewcommand{\thesection}{\Alph{section}}
\setcounter{section}{0}

\section{Overview}

We provide additional details in this appendix. In Sec. \ref{sec:supp-network}, we describe the details of our network architecture and implementation. In Sec. \ref{sec:supp-dataset}, we clarify our use of keywords for crawling background images. In Sec. \ref{sec:supp-training}, we explain how we train our model and show details of our data augmentations. In Sec. \ref{sec:supp-performance}, we show additional metrics about our method's performance. In Sec. \ref{sec:supp-results}, we show all the qualitative results used in our user study along with the average score per sample.
\section{Network}
\label{sec:supp-network}

\subsection{Architecture}
\textbf{Backbone}
Both ResNet and MobileNetV2 are adopted from the original implementation with minor modifications. We change the first convolution layer to accept 6 channels for both the input and the background images. We follow DeepLabV3's approach and change the last downsampling block with dilated convolutions to maintain an output stride of 16. We do not use the multi-grid dilation technique proposed in DeepLabV3 for simplicity.

\textbf{ASPP} We follow the original implementation of ASPP module proposed in DeepLabV3. Our experiment suggests that setting dilation rates to (3, 6, 9) produces the better results.

\textbf{Decoder}
\begin{center}
    \text{CBR128 - CBR64 - CBR48 - C37}
\end{center}

"CBR$k$" denotes $k$ 3$\times$3 convolution filters with same padding without bias followed by Batch Normalization and ReLU. "C$k$" denotes $k$ 3$\times$3 convolution filters with same padding and bias. Before every convolution, decoder uses bilinear upsampling with a scale factor of 2 and concatenates with the corresponding skip connection from the backbone. The 37-channel output consists of 1 channel of alpha $\alpha_c$, 3 channels of foreground residual $F^R_c$, 1 channel of error map $E_c$, and 32 channels of hidden features $H_c$. We clamp $\alpha_c$ and $E_c$ to 0 and 1. We apply ReLU on $H_c$.

\textbf{Refiner}

\begin{center}
    \begin{tabular}{ll}
        First stage: & C*BR24 - C*BR16 \\
        Second stage: & C*BR12 - C*4
    \end{tabular}
\end{center}

"C*BR$k$" and "C*$k$" follow the same definition above except that the convolution does not use padding.

Refiner first resamples coarse outputs $\alpha_c$, $F^R_c$, $H_c$, and input images $I$, $B$ to $\frac{1}{2}$ resolution and concatenates them as $[n\times42\times\frac{h}{2}\times\frac{w}{2}]$ features. Based on the error prediction $E_c$, we crop out top $k$ most error-prone patches $[nk\times42\times8\times8]$. After applying the first stage, the patch dimension becomes $[nk\times16\times4\times4]$. We upsample the patches with nearest upsampling and concatenate them with patches at the corresponding location from $I$ and $B$ to form $[nk\times22\times8\times8]$ features. After the second stage, the patch dimension becomes $[nk\times4\times4\times4]$. The 4 channels are alpha and foreground residual. Finally, we bilinearly upsample the coarse $\alpha_c$ and $F^R_c$ to full resolution and replace the refined patches to their corresponding location to form the final output $\alpha$ and $F^R$.

\subsection{Implementation}
We implement our network in PyTorch \cite{pytorch}. The patch extraction and replacement can be achieved via the native vectorized operations for maximum performance. We find that PyTorch's nearest upsampling operation is much faster on small-resolution patches than bilinear upsampling, so we use it when upsampling the patches.

\section{Dataset}
\label{sec:supp-dataset}

\begin{figure}[h!]
    \centering
    \includegraphics[width=0.45\textwidth]{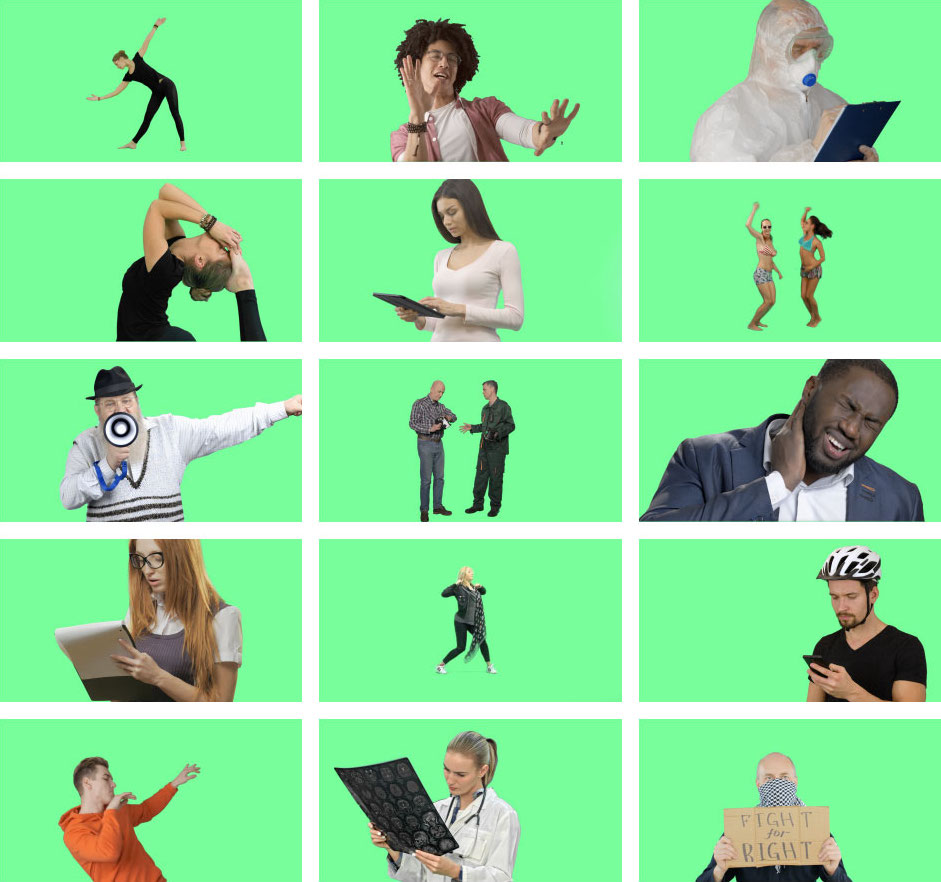}
    \caption{More examples from VideoMatte240K dataset.}
    \label{fig:supp-videomatte}
\end{figure}

\textbf{VideoMatte240K} The dataset contains 484 video clips, which consists a total of 240,709 frames. The average frames per clip is 497.3 and the median is 458.5. The longest clip has 1500 frames while the shortest clip has 124 frames. Figure \ref{fig:supp-videomatte} shows more examples from VideoMatte240K dataset.

\textbf{Background} The keywords we use for crawling background images are: 

\begin{scriptsize}
\begin{tt}
\setlength\tabcolsep{0.5 pt}
\vspace{2pt}
\hskip-1pt
\begin{tabular}{ccc}
    airport interior & attic & bar interior \\
    bathroom & beach & city \\
    church interior & classroom interior & empty city \\
    forest & garage interior & gym interior \\
    house outdoor & interior & kitchen \\
    lab interior & landscape & lecture hall \\
    mall interior & night club interior & office \\
    rainy woods & rooftop & stadium interior \\
    theater interior & train station & warehouse interior \\
    & workplace interior
\end{tabular}
\end{tt}
\end{scriptsize}
\section{Training}
\label{sec:supp-training}

Table \ref{tab:supp-training} records the training order, epochs, and hours of our final model on different datasets. We use 1$\times$RTX 2080 TI when training only the base network and 2$\times$RTX 2080 TI when training the network jointly.
\begin{table}[h!]
    \centering
    \setlength\tabcolsep{5.5 pt}
    \begin{tabularx}{.45\textwidth}{llrr}
        \toprule
        Dataset & Network & Epochs & Hours  \\
        \midrule
        VideoMatte240K & $G_{\text{base}}$ & 8 & 24 \\
        VideoMatte240K & $G_{\text{base}} + G_{\text{refine}}$ & 1 & 12 \\
        PhotoMatte13K & $G_{\text{base}} + G_{\text{refine}}$ & 25 & 8 \\
        Distinctions & $G_{\text{base}} + G_{\text{refine}}$ & 30 & 8 \\
        \bottomrule
    \end{tabularx}
    \caption{Training epochs and hours on different datasets. Time measured on model with ResNet-50 backbone.}
    \label{tab:supp-training}
\end{table}

Additionally, we use mixed precision training for faster computation and less memory consumption. When using multiple GPUs, we apply data parallelism to split the minibatch across multiple GPUs and switch to use PyTorch's Synchronized Batch Normalization to track batch statistics across GPUs.

\subsection{Training augmentation}

For every alpha and foreground training sample, we rotate to composite with backgrounds in a "zip" fashion to form a single epoch. For example, if there are 60 training samples and 100 background images, a single epoch is 100 images, where the 60 samples first pair with the first 60 background images, then the first 40 samples pair with the rest of the 40 background images again. The rotation stops when one set of images runs out. Because the datasets we use are very different in sizes, this strategy is used to generalize the concept of an epoch.

We apply random rotation ($\pm$5deg), scale (0.3$\sim$1), translation ($\pm$10\%), shearing ($\pm$5deg), brightness (0.85$\sim$1.15), contrast (0.85$\sim$1.15), saturation (0.85$\sim$1.15), hue ($\pm$0.05), gaussian noise ($\sigma^2\leq$0.03), box blurring, and sharpening independently to foreground and background on every sample. We then composite the input image using $I = \alpha F + (1 - \alpha)B$.

We additionally apply random rotation ($\pm$1deg), translation ($\pm$1\%), brightness (0.82$\sim$1.18), contrast (0.82$\sim$1.18), saturation (0.82$\sim$1.18), and hue ($\pm$0.1) only on the background 30\% of the time. This small misalignment between input $I$ and background $B$ increases model's robustness on real-life captures.

\begin{figure}[h!]
    \centering
    \includegraphics[width=0.45\textwidth]{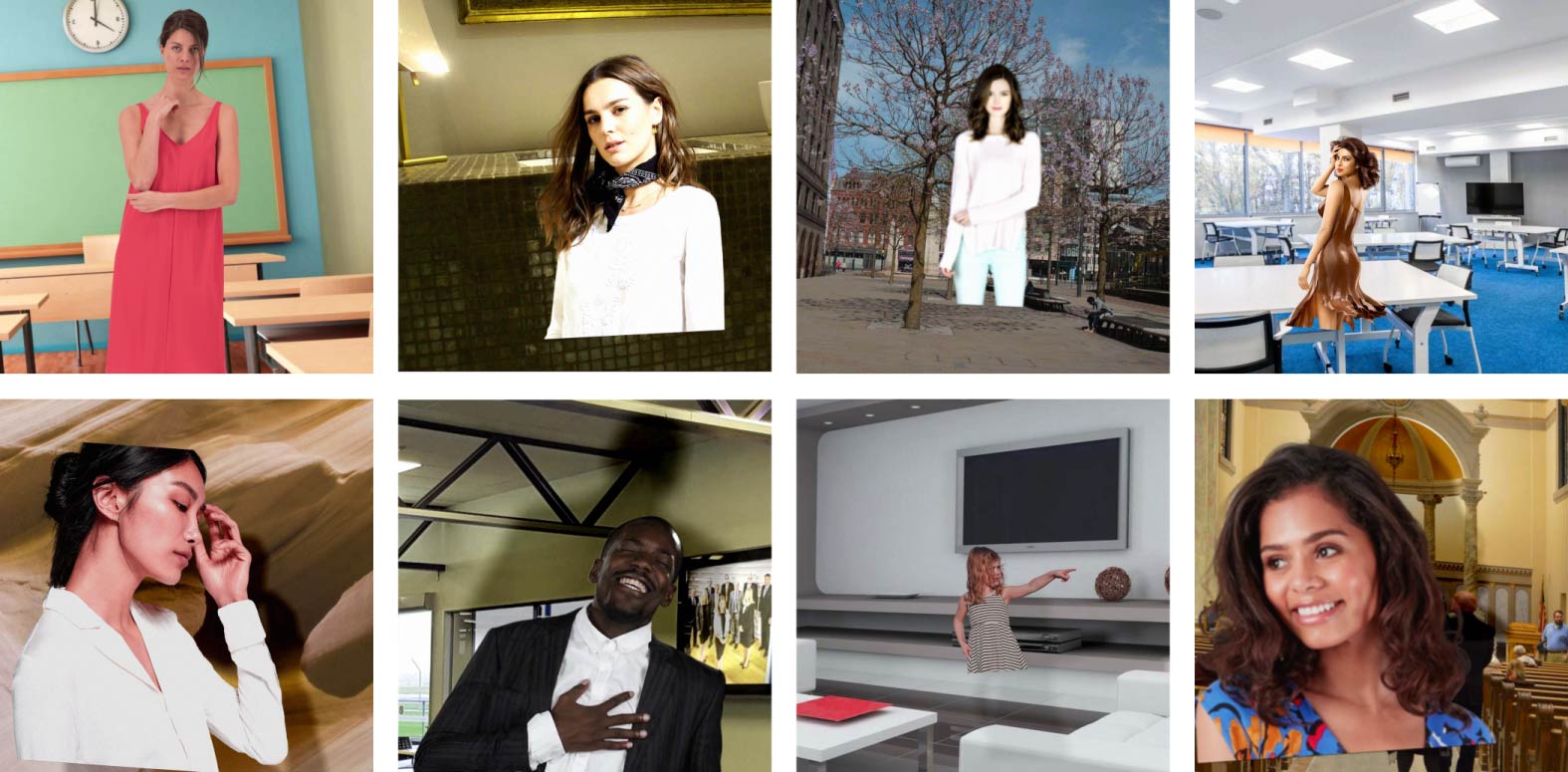}
    \caption{Training samples with augmentations. Bottom row are samples with shadow augmentation. Actual samples have different resolutions and aspect ratios.}
    \label{fig:supp-augmentation}
\end{figure}

We also find creating artificial shadows increases model's robustness because subjects in real-life often cast shadows on the environment. Shadows are created on $I$ by darkening some areas of the image behind the subject following the subject's contour 30\% of the time. Examples of composited images are shown in Figure \ref{fig:supp-augmentation}. The bottom row shows examples of shadow augmentation.

\subsection{Testing augmentation}

For AIM and Distinctions, which have 11 human test samples each, we pair every sample with 5 random backgrounds from the background test set. For PhotoMatte85, which has 85 test samples, we pair every sample with only 1 background. We use the method and metrics described in \cite{rhemann2009perceptually} to evaluate the resulting sets of 55, 55, and 85 images.

We apply a random subpixel translation ($\pm$0.3 pixels), random gamma (0.85$\sim$1.15), and gaussian noise ($\mu=\pm0.02$, $0.08\leq\sigma^2\leq0.15$) to background $B$ only, to simulate misalignment. 

The trimaps used as input for trimap-based methods and for defining the error metric regions are obtained by thresholding the grouth-truth alpha between 0.06 and 0.96, then applying 10 iterations of dilation followed by 10 iterations of erosion using a 3$\times$3 circular kernel.

\section{Performance}
\label{sec:supp-performance}

Table \ref{tab:supp-performance-gpus} shows the performance of our method on two Nvidia RTX 2000 series GPUs: the flagship RTX 2080 TI and the entry-level RTX 2060 Super. The entry-level GPU yields lower FPS but is still within an acceptable range for many real-time applications. Additionally, Table \ref{tab:supp-performance-batch-sizes} shows that switching to a larger batch size and a lower precision can increase the FPS significantly.


\begin{figure*}[h!]
    \centering
    \fbox{\includegraphics[width=0.95\textwidth]{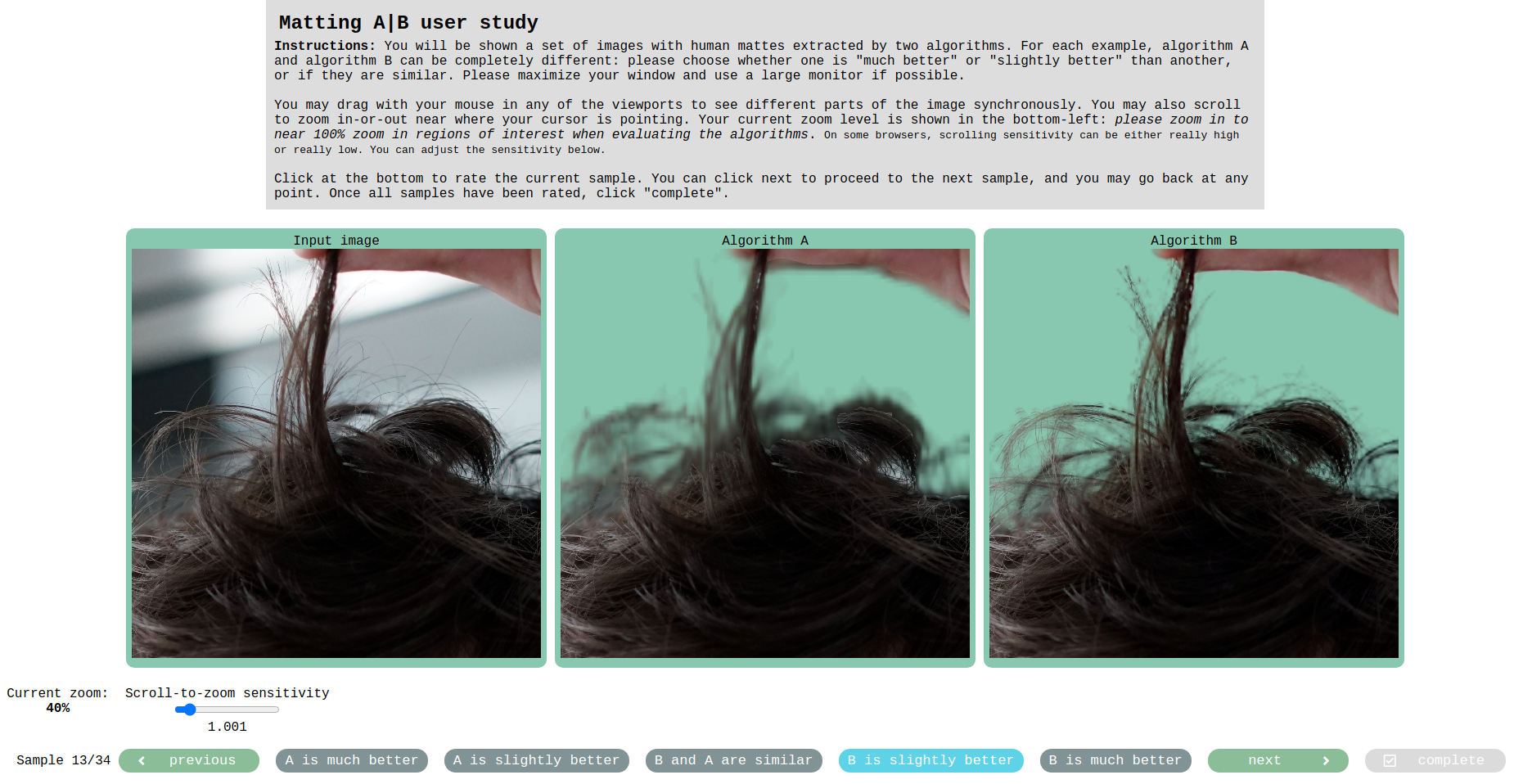}}
    \caption{The web UI for our user study. Users are shown the original image and two result images from Ours and BGM methods. Users are given the instruction to rate whether one algorithm is "much better", "slightly better", or both as "similar".}
    \label{fig:supp-ui}
\end{figure*}


\begin{table}[h!]
    \centering
    \setlength\tabcolsep{6.5 pt}
    \begin{tabularx}{.45\textwidth}{llcr}
        \toprule
        GPU & Backbone & Reso & FPS \\
        \midrule
        \multirow{4.5}{*}{RTX 2080 TI}
            & \multirow{2}{*}{ResNet-50}
                & HD & 60.0 \\
            &   & 4K & 33.2 \\
            \cmidrule{2-4}
            & \multirow{2}{*}{MobileNetV2}
                & HD & 100.6 \\
            &   & 4K & 45.4 \\
        \midrule
        \multirow{4.5}{*}{RTX 2060 Super}
            & \multirow{2}{*}{ResNet-50}
                & HD & 42.8 \\
            &   & 4K & 23.3 \\
            \cmidrule{2-4}
            & \multirow{2}{*}{MobileNetV2}
                & HD & 75.6 \\
            &   & 4K & 31.3 \\
        \bottomrule
    \end{tabularx}
    \caption{Performance on different GPUs. Measured with batch size 1 and FP32 precision.}
    \label{tab:supp-performance-gpus}
\end{table}

\begin{table}[h!]
    \centering
    \setlength\tabcolsep{6 pt}
    \begin{tabularx}{.45\textwidth}{lcccr}
        \toprule
        Backbone & Reso & Batch & Precision & FPS \\
        \midrule
        \multirow{4.5}{*}{MobileNetV2}
            & \multirow{3}{*}{HD}
                & 1 & FP32 & 100.6 \\
            &   & 8 & FP32 & 138.4 \\
            &   & 8 & FP16 & 200.0 \\
            \cmidrule{2-5}
            & 4K & 8 & FP16 & 64.2 \\
        \bottomrule 
    \end{tabularx}
    \caption{Performance using different batch sizes and precisions. Measured on RTX 2080 TI.}
    \label{tab:supp-performance-batch-sizes}
\end{table}

\section{Additional Results}
\label{sec:supp-results}

\begin{figure*}[h]
    \raggedleft
    \renewcommand{\arraystretch}{3.48}
    \setlength\tabcolsep{1 pt}
    \begin{large}
    \begin{tabularx}{.95\textwidth}{lX}
        \multirow{12}{*}{
            \includegraphics[width=0.882\textwidth]{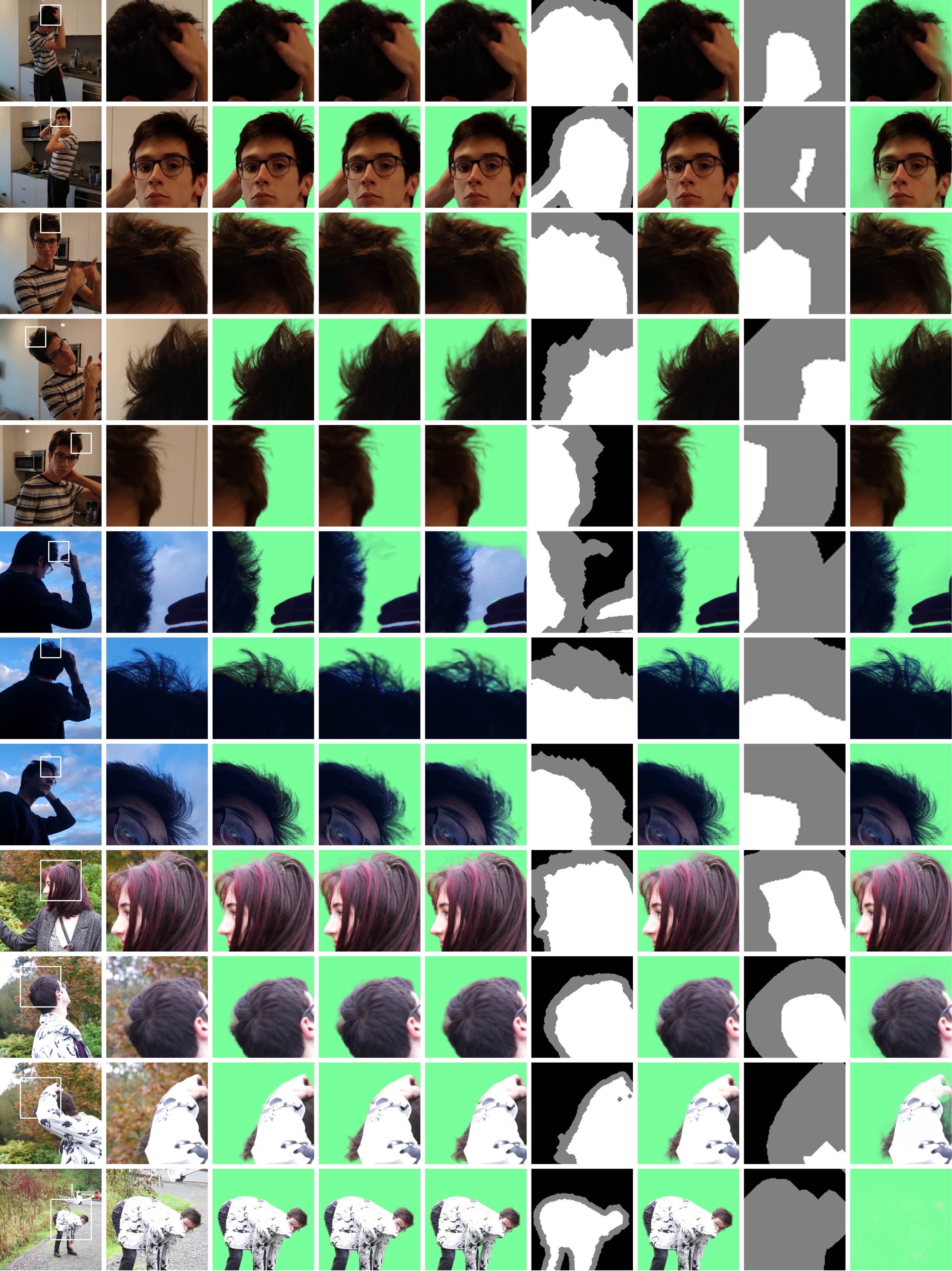}
        } & \textcolor{ForestGreen}{+2.0} \\
          & \textcolor{red}{-3.3} \\
          & \textcolor{ForestGreen}{+1.8} \\
          & \textcolor{ForestGreen}{+0.1} \\
          & \textcolor{ForestGreen}{+3.3} \\
          & \textcolor{ForestGreen}{+9.0} \\
          & \textcolor{ForestGreen}{+7.5} \\
          & \textcolor{ForestGreen}{+7.1} \\
          & \textcolor{red}{-2.6} \\
          & \textcolor{red}{-1.3} \\
          & \textcolor{red}{-0.8} \\
          & \textcolor{ForestGreen}{2.4}
    \end{tabularx}
    \end{large}
    \renewcommand{\arraystretch}{1}
    \setlength\tabcolsep{3 pt}
    \begin{small}
    \begin{tabularx}{.95\textwidth}{XX|XXX|XX|XX|l}
        & Input & Ours & BGM & BGM$_a$ & Trimap & FBA & Trimap & FBA$_{\text{auto}}$ & Score \\
        & & \multicolumn{3}{l|}{\footnotesize{Background-based methods}} & \multicolumn{2}{l|}{\footnotesize{Manual trimap}} & \multicolumn{2}{l|}{\footnotesize{Segmentation-morph trimap}} &
    \end{tabularx}
    \end{small}
    \caption{Additional qualitative comparison (1/3). Average user ratings between Ours and BGM are included. A score of -10 denotes BGM is "much better", -5 denotes BGM is "slightly better", 0 denotes "similar", +5 denotes Ours is "slightly better", +10 denotes Ours is "much better". Our method receives an average 3.1 score.}
    \label{fig:supp-qualitative-evaluation-1}
    \vspace{-1.5em}
\end{figure*}

\begin{figure*}[h]
    \raggedleft
    \renewcommand{\arraystretch}{3.48}
    \setlength\tabcolsep{1 pt}
    \begin{large}
    \begin{tabularx}{.95\textwidth}{lX}
        \multirow{11}{*}{
            \includegraphics[width=0.882\textwidth]{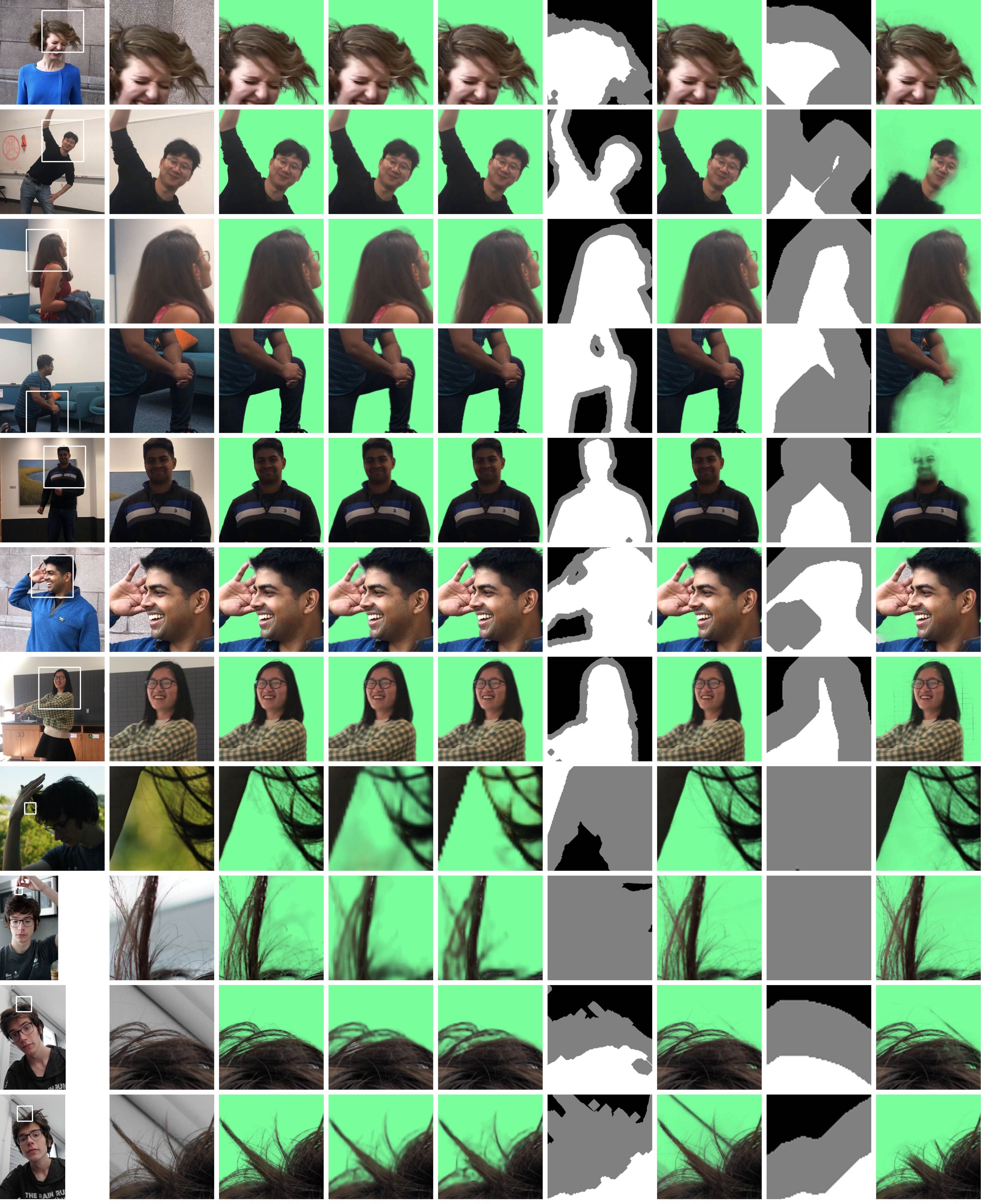}
        } & \textcolor{red}{-5.5} \\
          & \textcolor{ForestGreen}{+2.9} \\
          & \textcolor{ForestGreen}{+2.3} \\
          & \textcolor{ForestGreen}{+3.8} \\
          & \textcolor{ForestGreen}{+3.6} \\
          & \textcolor{red}{-3.6} \\
          & \textcolor{red}{-2.4} \\
          & \textcolor{ForestGreen}{+8.8} \\
          & \textcolor{ForestGreen}{+6.0} \\
          & \textcolor{ForestGreen}{+5.6} \\
          & \textcolor{ForestGreen}{+6.3} \\
    \end{tabularx}
    \end{large}
    \renewcommand{\arraystretch}{1}
    \setlength\tabcolsep{3 pt}
    \begin{small}
    \begin{tabularx}{.95\textwidth}{XX|XXX|XX|XX|l}
        & Input & Ours & BGM & BGM$_a$ & Trimap & FBA & Trimap & FBA$_{\text{auto}}$ & Score \\
        & & \multicolumn{3}{l|}{\footnotesize{Background-based methods}} & \multicolumn{2}{l|}{\footnotesize{Manual trimap}} & \multicolumn{2}{l|}{\footnotesize{Segmentation-morph trimap}} &
    \end{tabularx}
    \end{small}
    \caption{Additional qualitative comparisons (2/3)}
    \label{fig:supp-qualitative-evaluation-2}
    \vspace{-1.5em}
\end{figure*}

\begin{figure*}[h]
    \raggedleft
    \renewcommand{\arraystretch}{3.48}
    \setlength\tabcolsep{1 pt}
    \begin{large}
    \begin{tabularx}{.95\textwidth}{lX}
        \multirow{11}{*}{
            \includegraphics[width=0.882\textwidth]{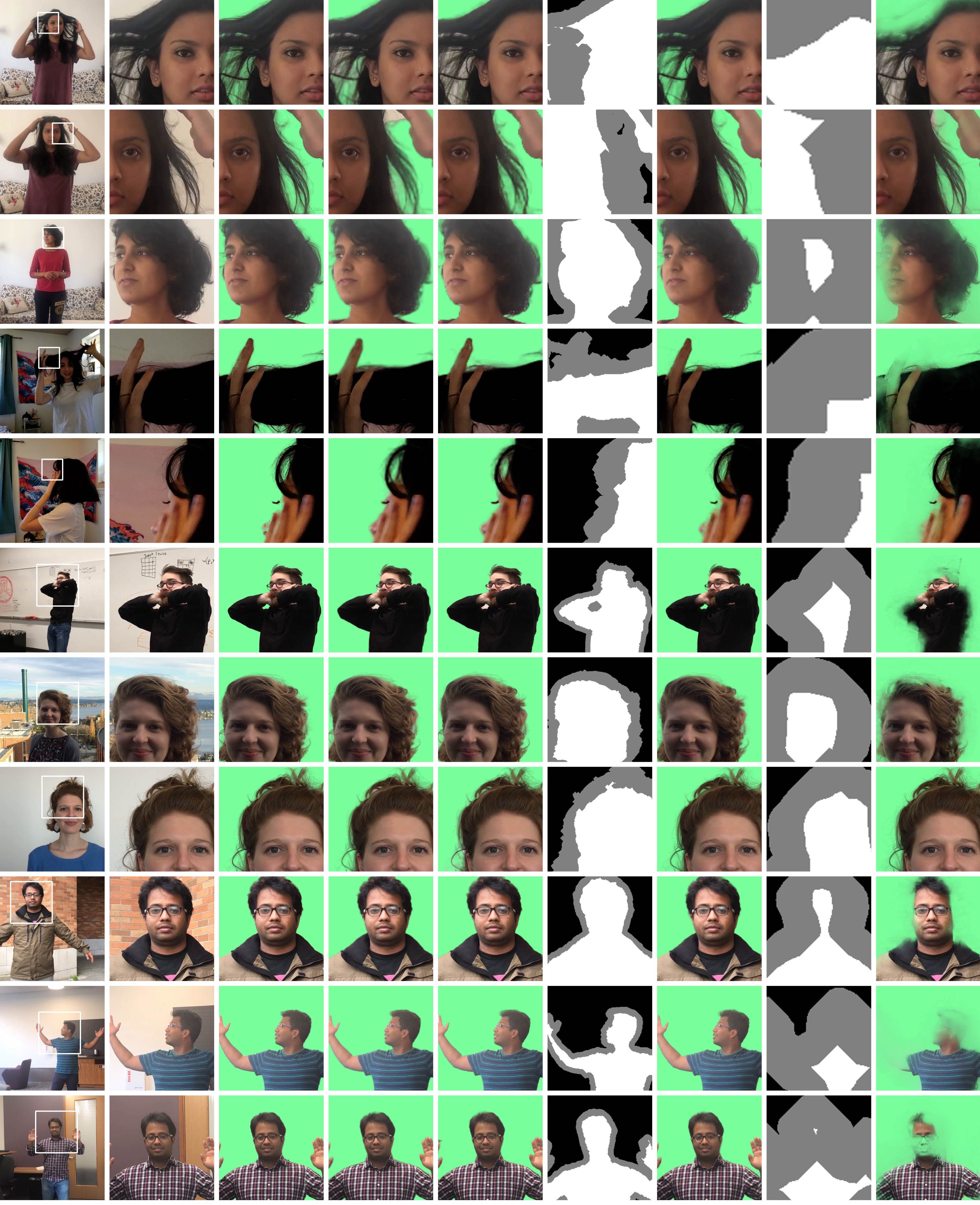}
        } & \textcolor{ForestGreen}{+7.4} \\
          & \textcolor{ForestGreen}{+8.8} \\
          & \textcolor{ForestGreen}{+7.4} \\
          & \textcolor{ForestGreen}{+7.0} \\
          & \textcolor{red}{-3.5} \\
          & \textcolor{ForestGreen}{+6.8} \\
          & \textcolor{red}{-3.3} \\
          & \textcolor{ForestGreen}{+5.8} \\
          & \textcolor{ForestGreen}{+5.1} \\
          & \textcolor{ForestGreen}{+0.1} \\
          & \textcolor{ForestGreen}{+3.8} \\
    \end{tabularx}
    \end{large}
    \renewcommand{\arraystretch}{1}
    \setlength\tabcolsep{3 pt}
    \begin{small}
    \begin{tabularx}{.95\textwidth}{XX|XXX|XX|XX|l}
        & Input & Ours & BGM & BGM$_a$ & Trimap & FBA & Trimap & FBA$_{\text{auto}}$ & Score \\
        & & \multicolumn{3}{l|}{\footnotesize{Background-based methods}} & \multicolumn{2}{l|}{\footnotesize{Manual trimap}} & \multicolumn{2}{l|}{\footnotesize{Segmentation-morph trimap}} &
    \end{tabularx}
    \end{small}
    \caption{Additional qualitative comparisons (3/3)}
    \label{fig:supp-qualitative-evaluation-3}
    \vspace{-1.5em}
\end{figure*}

In Figures \ref{fig:supp-qualitative-evaluation-1}, \ref{fig:supp-qualitative-evaluation-2}, \ref{fig:supp-qualitative-evaluation-3}, we show all 34 examples in the user study, along with their average rating and results by different methods. Figure \ref{fig:supp-ui} shows the web UI for our user-study. 

\end{document}